\documentclass[letterpaper]{article} 
\usepackage{aaai2026}  
\usepackage{times}  
\usepackage{helvet}  
\usepackage{courier}  
\usepackage[hyphens]{url}  
\usepackage{graphicx} 
\usepackage{natbib}  
\usepackage{caption} 
\usepackage{algorithm}
\usepackage{algorithmic}

\usepackage{newfloat}
\usepackage{listings}
\DeclareCaptionStyle{ruled}{labelfont=normalfont,labelsep=colon,strut=off} 
\floatstyle{ruled}
\newfloat{listing}{tb}{lst}{}
\floatname{listing}{Listing}

\usepackage{bm}
\usepackage{amsmath}
\usepackage{amssymb}
\usepackage{booktabs}
\usepackage{multirow}
\usepackage{pifont}
\usepackage{subcaption}
\newcommand{\xmark}{\ding{55}}
\newcommand{\cmark}{\ding{51}}
\newcommand\blfootnote[1]{
  \begingroup
  \renewcommand\thefootnote{}\footnote{#1}
  \addtocounter{footnote}{-1}
  \endgroup
}
\newcommand{\cam}{\textcolor{black}}

\title{CANDI: Curated Test-Time Adaptation for Multivariate Time-Series \\ Anomaly Detection Under Distribution Shift}
\author{
    HyunGi Kim\textsuperscript{\rm 1}, 
    Jisoo Mok\textsuperscript{\rm 2}, 
    Hyungyu Lee\textsuperscript{\rm 1}, 
    Juhyeon Shin\textsuperscript{\rm 3}, 
    Sungroh Yoon\textsuperscript{\rm 1, \rm 3, \rm 4 \dag}
}
\affiliations{
    \textsuperscript{\rm 1}Department of ECE, Seoul National University\\
    \textsuperscript{\rm 2}DGIST, 
    \textsuperscript{\rm 3}IPAI, Seoul National University, 
    \textsuperscript{\rm 4}AIIS, ASRI, and INMC, Seoul National University\\

    rlagusrl0128@snu.ac.kr, jmok@dgist.ac.kr, \{rucy74, newjh12, sryoon\}@snu.ac.kr
}

\usepackage{bibentry}

\begin{document}

\maketitle

\begin{abstract}
Multivariate time-series anomaly detection (MTSAD) aims to identify deviations from normality in multivariate time-series and is critical in real-world applications. However, in real-world deployments, distribution shifts are ubiquitous and cause severe performance degradation in pre-trained anomaly detector. Test-time adaptation (TTA) updates a pre-trained model on-the-fly using only unlabeled test data, making it promising for addressing this challenge. In this study, we propose \textbf{CANDI} (\textbf{C}urated test-time adaptation for multivariate time-series \textbf{AN}omaly detection under \textbf{DI}stribution shift), a novel TTA framework that selectively identifies and adapts to potential false positives while preserving pre-trained knowledge. CANDI introduces a False Positive Mining (FPM) strategy to curate adaptation samples based on anomaly scores and latent similarity, and incorporates a plug-and-play Spatiotemporally-Aware Normality Adaptation (SANA) module for structurally informed model updates. Extensive experiments demonstrate that CANDI significantly improves the performance of MTSAD under distribution shift, improving AUROC up to 14\% while using fewer adaptation samples.
\end{abstract}

\begin{links}
    \link{Code}{https://github.com/kimanki/CANDI}
\end{links}

\blfootnote{$^\dag$ Corresponding Author\\}

\section{Introduction}
Multivariate time-series anomaly detection (MTSAD) aims to identify abnormal patterns within multivariate time-series data, which contain multiple interdependent variables~\cite{wang2025survey, li2023deep, choi2021deep}.
This capability is essential for maintaining the stability, safety, and efficiency of complex real-world systems through continuous monitoring of their states and timely decision-making~\cite{app14167055, shin2020itad}.
Accurate and robust MTSAD models are thus crucial for ensuring the stable operation of high-stakes, time-sensitive applications, such as industrial maintenance~\cite{tanuska2021smart} and healthcare monitoring~\cite{galvao2024anomaly}. 

Due to the scarcity of labeled anomaly data in real-world scenarios, most MSTAD methods adopt unsupervised approaches~\cite{AnomalyTransformer, MEMTO, CAROTS}. 
These methods generally assume that the training data consists only of normal operating conditions and learn to model normality. 
One common approach is reconstruction-based anomaly detection~\cite{wu2025catch, AnomalyTransformer, MSCRED}.
Here, the model, typically in the form of an autoencoder, is trained to reconstruct normal time-series data, and at test-time, samples that yield high reconstruction errors are identified as anomalies. 
Density-based methods estimate the probability density of normal time-series and flag data with low likelihoods as anomalies~\cite{MTGFlow, dai2022graphaugmented}. 
Lastly, distance-based approaches detect anomalies by measuring the distance of test data to normal clusters in a learned embedding space~\cite{THOC, CTAD}. 

\begin{figure}[t]
    \centering
    \includegraphics[width=\linewidth]{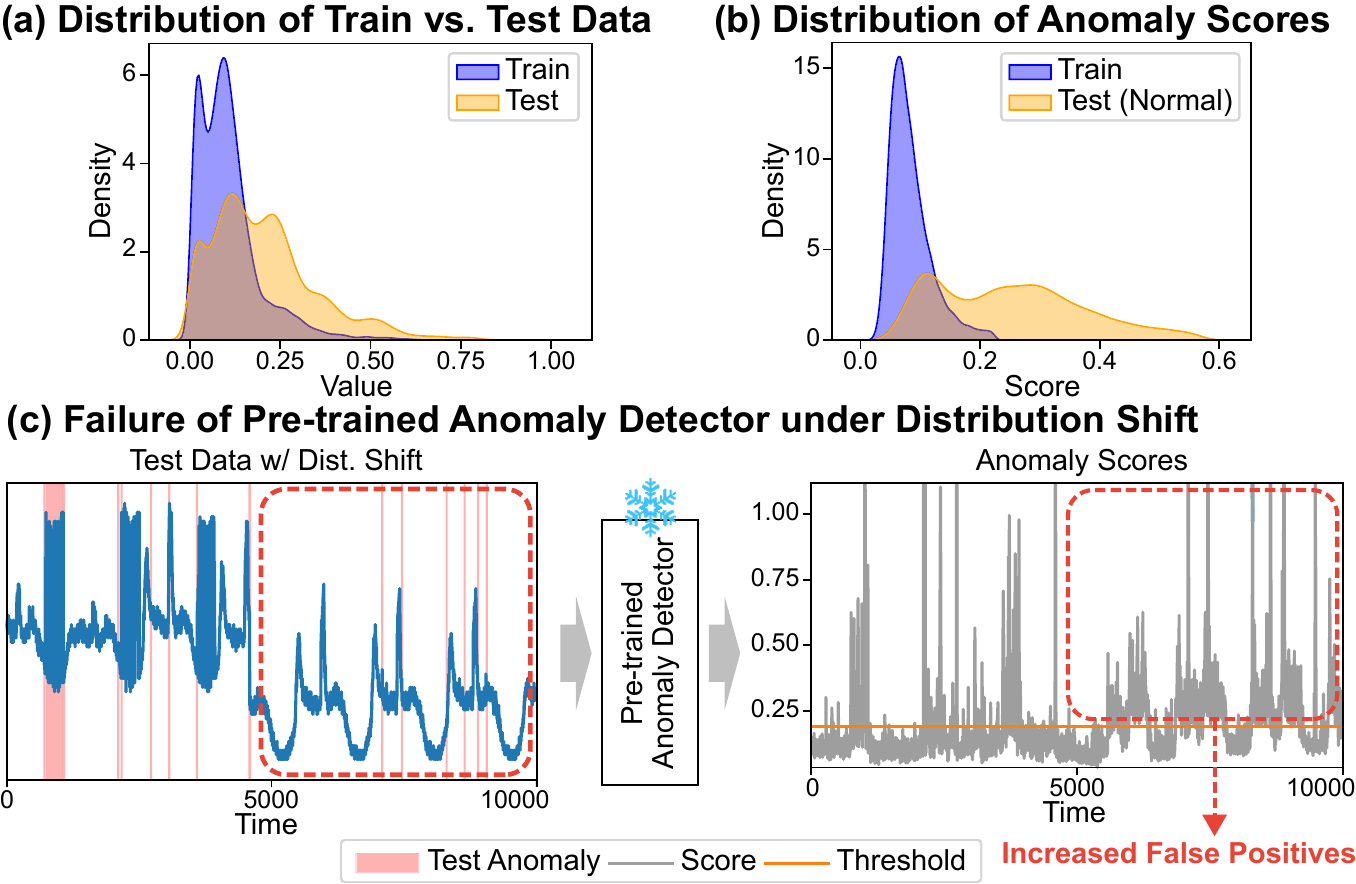}
    \caption{[Top] Real-world time-series data often exhibit non-stationarity, leading to continuous distribution shifts between training and test data. [Bottom] As shown in the later part of the anomaly scores, under distribution shift, pre-trained anomaly detectors can provide excessive false positives, undermining reliability under deployment.}
    \label{fig:motivation}
    \vspace{-10pt}
\end{figure}

Despite recent advances, most MTSAD approaches assume that the training and test data belong to the same distribution.
However, this assumption is often violated in real-world systems, due to numerous factors that cause a shift in the data distribution,~\textit{e.g.,} changes in system dynamics, sensor drifts, or environmental changes~\cite{karimi2010extensive}. 
Such distribution shifts induce normality shifts, where previously unseen but normal patterns emerge in the test data~\cite{han2023anomaly}. 
As shown in Figure~\ref{fig:motivation}, MTSAD models that have not been adapted to these shifts are prone to misclassifying these new normal patterns as anomalies, leading to a substantial increase in false positives. 

Test-time adaptation (TTA) refers to the paradigm of adapting a pre-trained model at inference time using only unlabeled test data~\cite{wang2020tent}. 
While TTA has shown success in tasks such as image classification and segmentation~\cite{lee2024entropy, gao2023atta}, its application to MTSAD remains largely underexplored. 
In MTSAD, TTA offers a promising avenue towards addressing continuously evolving distribution. 
A prior work~\cite{kim2024model} performs TTA on MTSAD by updating all trainable parameters on test samples that are identified as normal. 
This approach suffers from two major setbacks.
First, it completely disregards false positives, \textit{i.e.}, normal samples that are misclassified as anomalies.
These false positives can provide informative learning signal as they correspond to underrepresented normal patterns, revealing regions where the model needs further adaptation.
Second, adaptation of all trainable parameters may overwrite useful representations learned during pre-training.

To address the challenges of MTSAD posed by distribution shift, we propose \textbf{CANDI} (\textbf{C}urated test-time adaptation for multivariate time-series \textbf{AN}omaly detection under \textbf{DI}stribution shift), a novel TTA framework for MTSAD that adapts a pre-trained anomaly detector by curating informative test samples while preserving the original knowledge of the detector. 
CANDI is built on a reconstruction-based anomaly detector and introduces two key components: False Positive Mining (FPM) and Spatiotemporally-Aware Normality Adaptation (SANA). 
FPM identifies potential false positives based on their anomaly scores and proximity in latent space to normal validation samples. 
These challenging-to-detect yet reliable samples are used for adaptation. 

Instead of updating the entire model, SANA provides a lightweight, plug-and-play adaptation module that captures temporal and inter-variable shifts via temporal convolutions and an attention mechanism, while keeping the backbone frozen. 
By combining selective adaptation signals with a safe adaptation mechanism, CANDI enhances robustness and accuracy under distribution shift without compromising the pre-trained model. 
Through extensive experiments, CANDI demonstrates significant gains over baselines under distribution shifts, including a 14\% improvement of AUROC compared to the TTA baseline despite using less than 2\% of the total test data for adaptation.

In summary, our contributions are as follows:
\begin{itemize}
  \item We identify and address the critical challenge of distribution shift in MTSAD, a problem that causes significant false positives in real-world systems.
  \item We propose CANDI, a novel TTA framework for MTSAD that curates informative samples via false positive mining, and adapts the model with a spatiotemporally-aware module while preserving pre-trained knowledge.
  \item We demonstrate that CANDI consistently outperforms MTSAD baselines, achieving up to a 14\% AUROC gain while using less than 2\% of the data for adaptation.
\end{itemize}

\begin{figure*}[t]
    \centering
    \includegraphics[width=\linewidth]{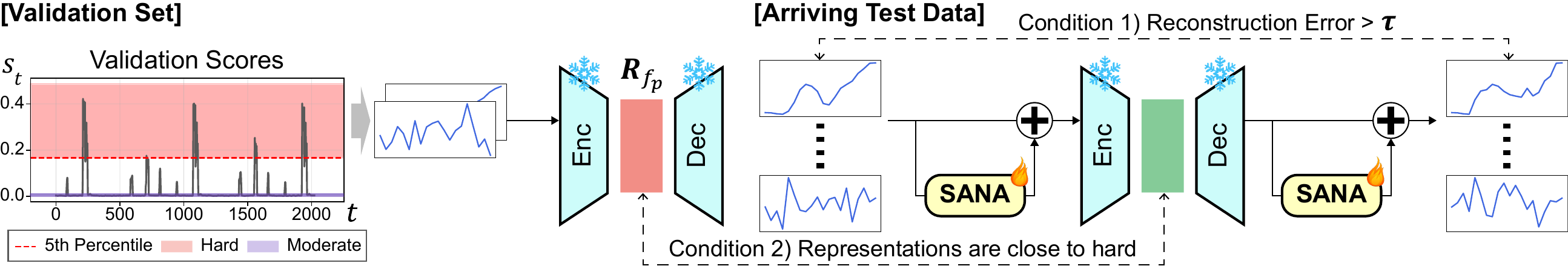}
    \caption{Overall framework of CANDI. [Left] Anomaly scores are first computed on a normal validation set, and latent representations of samples falling within the top $\alpha$-percentile (\textit{e.g.}, 5th percentile) are extracted and stored in a reference false positive set $\bm{\mathcal{R}}_{fp}$. [Right] For arriving test data, if the anomaly score is above the threshold, its latent representation is compared to those in $\bm{\mathcal{R}}_{fp}$. If the distance is sufficiently small, the sample is identified as a potential false positive and used for adaptation. Adaptation is performed via the plug-and-play \textit{Spatiotemporally-Aware Normality Adaptation} (SANA) module, which updates only a lightweight residual component while preserving the knowledge and latent space of the pre-trained anomaly detector.}
    \label{fig:overview}
\end{figure*}

\section{Related Works}
\subsection{Unsupervised Multivariate Time-series Anomaly Detection}
Unsupervised multivariate time-series anomaly detection (MTSAD) has been studied across diverse paradigms.
Traditional methods like LOF~\cite{breunig2000lof} and one-class SVM~\cite{manevitz2001one} have been applied, but often fail to capture temporal dependencies.
Recent deep models fall into reconstruction-, density-, and graph-based categories.
Reconstruction-based models detect anomalies based on reconstruction errors~\cite{wu2025catch, AnomalyTransformer}.
USAD~\cite{USAD} extends this approach by introducing adversarial training between dual decoders.
Density-based models such as OmniAnomaly~\cite{SMD} use variational autoencoders to detect low-likelihood patterns.
Structure-aware models focus on inter-variable relations: MSCRED~\cite{MSCRED} reconstructs multi-scale correlation maps, GDN~\cite{deng2021graph} applies graph neural networks, and TimesNet~\cite{wu2022timesnet} leverages frequency-aware blocks for improved detection.

However, most unsupervised MTSAD models assume a static normal distribution after training. 
In reality, normality may drift due to system aging, sensor noise, or environmental changes~\cite{METER, liu2023anomaly, han2023anomaly}.
Without adaptation, false positives increase over time. 
Our work addresses this by enabling test-time refinement using unlabeled but selectively informative samples.

\subsection{Test-time Adaptation}
Test-time adaptation (TTA)~\cite{liang2025comprehensive, wang2020tent, niu2023towards, jia2024tinytta} is a paradigm that updates a pre-trained model at inference time to address distribution shifts, using only unlabeled test data.
In the domain of image classification, TTA methods, such as TENT~\cite{wang2020tent} and MEMO~\cite{zhang2022memo}, adjust model parameters by minimizing prediction entropy or self-supervised losses. 
More recent frameworks like CoTTA~\cite{wang2022continual} further explore the continuous adaptation of a pre-trained image classifier while addressing the risk of catastrophic forgetting.

Due to the ever-evolving, dynamic nature of real-world time-series data, extending TTA to time-series data is a natural yet under-explored direction~\cite{TAFAS}. 
TTA allows models to track evolving normal distribution and maintain performance under non-stationary conditions. 
For instance, M2N2~\cite{kim2024model} proposes to adaptively detrend input signals and update model parameters using test samples predicted as normal, demonstrating the potential of TTA in MTSAD.

However, the existing TTA approach for MTSAD considers normality shift narrowly, focusing primarily on changes in overall trends while overlooking more complex temporal and inter-variable distribution shifts. 
Their reliance on limited adaptation cues and the practice of updating the full model can increase the risk of catastrophic forgetting by overwriting pre-trained knowledge. 
Furthermore, they overlook the adaptation potential of difficult-to-detect yet informative false positive samples. 

\section{CANDI: Curated Test-time Adaptation for Multivariate Time-series Anomaly Detection}
In this section, we present \textbf{CANDI}, a TTA framework for MTSAD. As illustrated in Figure~\ref{fig:overview}, CANDI selectively adapts a pre-trained anomaly detector to curated informative test samples while preserving the pre-trained knowledge. It is comprised of two components: (1) \textit{False Positive Mining} that selects potential false positives based on anomaly score distribution and similarity in the latent space, and (2) \textit{Spatiotemporally-Aware Normality Adaptation} module that handles test-time distribution shifts in temporal and inter-variable patterns.

\subsection{Problem Formulation}
Let $\bm{X} = [\bm{x}_1, \bm{x}_2, \dots, \bm{x}_T] \in \mathbb{R}^{D \times T}$ denote a multivariate time-series with $D$ variables over $T$ time steps, where $\bm{x}_t \in \mathbb{R}^D$ is the observation at time $t$. The goal of MTSAD is to detect time steps or segments that deviate from the normal distribution.
The train, validation, and test data are obtained by splitting $\bm{X}$ into contiguous segments in chronological order. Following the standard assumption in unsupervised MTSAD, the train and validation data consist solely of normal, while the test data include anomalies to be detected.

CANDI adopts a reconstruction-based approach, in which a pre-trained anomaly detector $f_{\bm{\theta}}$ is an autoencoder trained to reconstruct a sliding input window of length $L$:
\begin{equation}
\hat{\bm{X}}_{t-L+1:t} = f_{\bm{\theta}}(\bm{X}_{t-L+1:t}).
\end{equation}

At inference time, the anomaly score $s_t$ is computed as the average squared reconstruction error:
\begin{equation}
s_t = \frac{1}{D \cdot L} \left\| \bm{X}_{t-L+1:t} - \hat{\bm{X}}_{t-L+1:t} \right\|_2^2,
\end{equation}
where higher scores indicate potential anomalies. However, in real-world deployments, the test distribution may shift due to factors like sensor drift or changing system dynamics. Our goal is to adapt the model to such shifts at test-time, without relying on labeled data or retraining the full model.

\subsection{False Positive Mining}
\label{sec:sample-mining}
Rather than adapting to all test samples with low anomaly scores, we selectively identify samples that cannot easily be detected and thus are likely to contribute meaningfully to adaptation. 
This curated sample selection not only mitigates the risk of performance degradation by avoiding unreliable samples for adaptation, but also improves adaptation efficiency by reducing the number of test samples used. 
In particular, we focus on samples that are challenging for the model to detect, such as potential false positives that reflect ambiguous or underrepresented normality. 
In consequence, CANDI focuses on areas where the model's prediction is uncertain, improving robustness with fewer updates.

We first compute anomaly scores for all samples in a validation set that contains only normal data. Let $\bm{\mathcal{S}}_{\text{val}} = \{s_i^{\text{val}}\}_{i=1}^{N_{\text{val}}}$ denote the set of anomaly scores computed on this validation set, where $s_i^{\text{val}}$ is the anomaly score for the $i$-th validation sample and $N_{\text{val}}$ is the total number of samples in the validation set. We then define a threshold $\tau$ as the $\alpha$-percentile of this score set:
\begin{equation}
\tau = \text{Percentile}(\bm{\mathcal{S}}_{\text{val}}, \alpha).
\end{equation}

Following standard practice, test samples with \(s_t > \tau\) are initially considered to be anomalous. 
However, since some normal samples in the validation set also exceed the same threshold \(\tau\) (\textit{i.e.}, \(s_i^{\text{val}} > \tau\)), we hypothesize that a subset of high-scoring test samples may likewise be false positives—normal but difficult instances that the model failed to capture during training. 
To identify these false positives, we collect validation samples with \(s_i^{\text{val}} > \tau\) and extract their latent representations using the frozen pre-trained encoder: \(\bm{z}_i = f^{\text{enc}}(\bm{X}_i^{\text{val}})\). We aggregate these into a reference set of false positive samples, denoted by \(\bm{\mathcal{R}}_{\text{fp}}\).

These reference samples reflect normal instances that exhibit unexpectedly high anomaly scores, suggesting that they lie near the decision boundary and share latent features with difficult-to-classify cases. 
To identify potential test-time false positives, we measure their proximity to known false positives from the validation set in the latent space using Mahalanobis distance. 
To provide stability, we estimate the mean \(\bm{\mu}\) and covariance matrix \(\bm{\Sigma}\) from the full set of latent representations of validation samples:
\begin{equation}
    \bm{\mu}, \bm{\Sigma} = \text{MeanCov}(\bm{\mathcal{R}}_{\text{val}}), \quad \bm{\mathcal{R}}_{\text{val}} = \{ f^{\text{enc}}(\bm{X}_i^{\text{val}}) \}_{i=1}^{N_{\text{val}}}.
\end{equation}
Using the full normal validation set ensures that the latent distance is measured with respect to the overall distribution of normal patterns, providing robustness and avoiding bias from sparsely sampled or ambiguous subsets like \(\bm{\mathcal{R}}_{\text{fp}}\).

For each test sample predicted as anomalous (\(s_t > \tau\)), we compute its latent representation \(\bm{z}_t = f^{\text{enc}}(\bm{X}_{t-L+1:t})\) and calculate its minimum squared Mahalanobis distance to the reference set \(\bm{\mathcal{R}}_{\text{fp}}\):
\begin{equation}
\mathcal{D}_M^2(\bm{z}_t, \bm{\mathcal{R}}_{\text{fp}}) = \min_{\bm{z}_r \in \bm{\mathcal{R}}_{\text{fp}}} (\bm{z}_t - \bm{z}_r)^\top \bm{\Sigma}^{-1} (\bm{z}_t - \bm{z}_r).
\end{equation}
We consider \(\bm{X}_{t-L+1:t}\) a potential false positive and include it for adaptation if this distance is below the 5th percentile of the chi-squared distribution with latent dimension \(d\). The threshold is defined as:
\begin{equation}
\delta = F^{-1}_{\chi^2_d}(0.05),
\end{equation}
where \(F^{-1}_{\chi^2_d}(\cdot)\) denotes the inverse cumulative distribution function. The inclusion criterion becomes:
\begin{equation}
\mathcal{D}_M^2(\bm{z}_t, \bm{\mathcal{R}}_{\text{fp}}) < \delta.
\end{equation}

This thresholding strategy is grounded in the statistical property that the squared Mahalanobis distance follows a chi-squared distribution with \(d\) degrees of freedom when latent representations of normal samples approximately follow a multivariate Gaussian. Thus, \(\delta\) defines a tight neighborhood around the reference set in latent space, and selecting the 5th percentile ensures only test samples with representations sufficiently close to those of high-scoring validation samples are selected. These samples likely share subtle but informative patterns, making them suitable candidates for adaptation. The adaptation set $\bm{\mathcal{A}}$ is constructed as:
\begin{equation}
    \bm{\mathcal{A}} = \left\{ \bm{X}_{t-L+1:t} \,\middle|\, 
    \left( s_t > \tau \right) \ \land\ 
    \left( \mathcal{D}_M^2\big(\bm{z}_t, \bm{\mathcal{R}}_{\mathrm{fp}}\big) < \delta \right) 
    \right\}.
\end{equation}

To further improve robustness, we also incorporate predicted normal samples with moderately high anomaly scores into the adaptation process. Specifically, we identify validation samples whose scores fall within the interquartile range ($Q_1$--$Q_3$) and store their latent representations as a separate reference set $\bm{\mathcal{R}}_{\text{mod}}$. These samples are not clearly anomalous but deviate enough from typical patterns to indicate areas where the model’s understanding of normality may be incomplete. For each test sample whose anomaly score is smaller than the threshold $\tau$, we compute its squared Mahalanobis distance to $\bm{\mathcal{R}}_{\text{mod}}$ using the same criterion as before. If the distance falls below the threshold $\delta$, the sample is included in the final adaptation set $\bm{\mathcal{A}}$.

\begin{table*}[t]
\footnotesize
\centering
\begin{tabular}{l|l|ccc|ccc|ccc}
\toprule
\multirow{2}{*}{\textbf{Dataset}} & \multirow{2}{*}{\textbf{Metric}} & \multicolumn{3}{c|}{$\alpha = 0.5\%$} & \multicolumn{3}{c|}{$\alpha = 1.0\%$} & \multicolumn{3}{c}{$\alpha = 5.0\%$} \\
& & Pretrained & M2N2 & CANDI & Pretrained & M2N2 & CANDI & Pretrained & M2N2 & CANDI \\
\midrule
\multirow{3}{*}{SWaT}
& AUROC & 0.827 & \textbf{0.891} & 0.889 & 0.827 & \textbf{0.891} & 0.890 & 0.827 & \textbf{0.891} & 0.888 \\
& AUPRC & 0.719 & 0.771 & \textbf{0.779} & 0.719 & 0.771 & \textbf{0.780} & 0.719 & 0.772 & \textbf{0.781} \\
& F1     & 0.291 & 0.711 & \textbf{0.752} & 0.287 & 0.700 & \textbf{0.738} & 0.291 & \textbf{0.636} & 0.624 \\
\midrule
\multirow{3}{*}{SMD\_1-7}
& AUROC & 0.883 & 0.901 & \textbf{0.922} & 0.883 & 0.864 & \textbf{0.923} & 0.883 & 0.886 & \textbf{0.922} \\
& AUPRC & 0.703 & 0.728 & \textbf{0.736} & 0.703 & 0.734 & \textbf{0.737} & 0.703 & 0.718 & \textbf{0.737} \\
& F1     & 0.103 & \textbf{0.662} & 0.107 & 0.562 & \textbf{0.707} & 0.688 & 0.724 & 0.723 & \textbf{0.725} \\
\midrule
\multirow{3}{*}{SMD\_1-8}
& AUROC & 0.719 & 0.837 & \textbf{0.872} & 0.719 & 0.805 & \textbf{0.872} & 0.719 & 0.772 & \textbf{0.867} \\
& AUPRC & 0.332 & 0.407 & \textbf{0.432} & 0.332 & 0.376 & \textbf{0.434} & 0.332 & 0.354 & \textbf{0.423} \\
& F1     & 0.377 & \textbf{0.406} & 0.393 & 0.362 & 0.389 & \textbf{0.409} & 0.092 & 0.115 & \textbf{0.213} \\
\midrule
\multirow{3}{*}{SMD\_2-1}
& AUROC & 0.648 & 0.698 & \textbf{0.725} & 0.648 & 0.693 & \textbf{0.711} & 0.648 & 0.683 & \textbf{0.780} \\
& AUPRC & 0.275 & 0.307 & \textbf{0.319} & 0.275 & 0.302 & \textbf{0.314} & 0.275 & 0.296 & \textbf{0.348} \\
& F1     & 0.265 & 0.266 & 0.266 & \textbf{0.296} & 0.295 & 0.292 & 0.273 & 0.309 & \textbf{0.327} \\
\midrule
\multirow{3}{*}{SMD\_2-4}
& AUROC & 0.821 & 0.895 & \textbf{0.908} & 0.821 & 0.895 & \textbf{0.908} & 0.821 & 0.828 & \textbf{0.899} \\
& AUPRC & 0.457 & 0.605 & \textbf{0.608} & 0.457 & 0.605 & \textbf{0.608} & 0.457 & 0.461 & \textbf{0.600} \\
& F1     & \textbf{0.357} & 0.355 & 0.352 & \textbf{0.378} & 0.377 & 0.372 & 0.316 & 0.311 & \textbf{0.512} \\
\midrule
\multirow{3}{*}{SMD\_3-2}
& AUROC & 0.451 & 0.573 & \textbf{0.717} & 0.451 & 0.632 & \textbf{0.717} & 0.451 & 0.640 & \textbf{0.717} \\
& AUPRC & 0.159 & 0.174 & \textbf{0.199} & 0.159 & 0.179 & \textbf{0.199} & 0.159 & 0.188 & \textbf{0.199} \\
& F1     & 0.017 & 0.017 & 0.017 & \textbf{0.031} & 0.030 & 0.030 & 0.247 & 0.194 & \textbf{0.262} \\
\bottomrule
\end{tabular}
\caption{Performance of test-time adaptation methods for multivariate time-series anomaly detection under test-time distribution shift. Bold denotes the best result for each metric and threshold level.
Each threshold is determined by the $\alpha$-percentile of validation anomaly scores.}
\label{tab:main-results}
\end{table*}

\subsection{Spatiotemporally-Aware Normality Adaptation}
\label{sec:adaptation-module}
\begin{figure}[t]
    \centering
    \includegraphics[width=\linewidth]{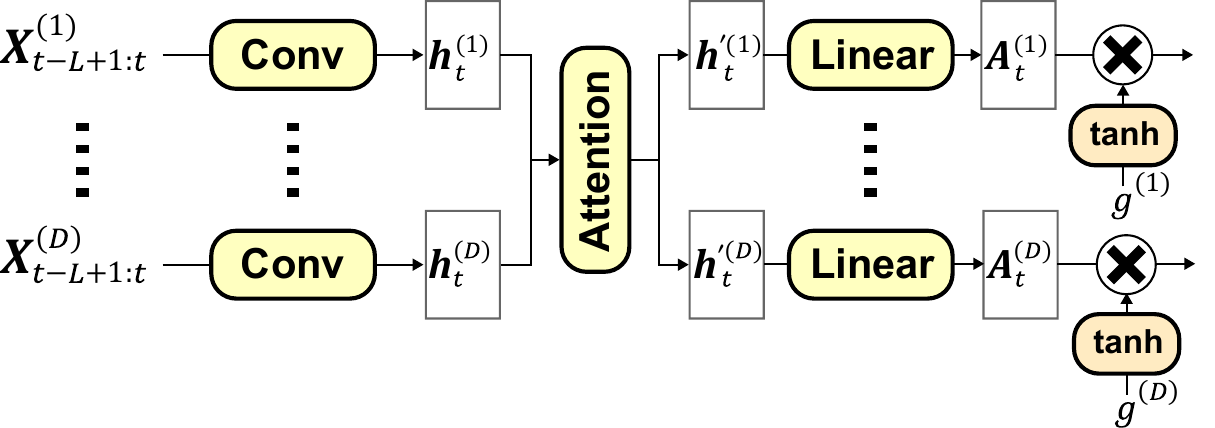}
    \caption{Architecture of the Spatiotemporally-Aware Normality Adaptation (SANA) module.}
    \label{fig:sana_overview}
    \vspace{-10pt}
\end{figure}

To enable stable and efficient TTA while preserving the knowledge of a pre-trained detector, we introduce a lightweight plug-and-play normality adaptation module, as illustrated in Figure~\ref{fig:sana_overview}. Motivated by TAFAS~\cite{TAFAS}, the module is attached to both the input and output of a frozen reconstruction-based anomaly detector. \cam{However, unlike TAFAS, which uses independent per-variable simple linear layers, we design a spatiotemporally-aware module composed of temporal convolution and inter-variable attention~\cite{liu2024itransformer}.} This allows our approach to capture distributional shifts occurring not only within each variable but also across variables through their interactions, providing a more expressive adaptation.

The input-side normality adaptation module adjusts incoming test samples to better align with the training-time normality distribution, enabling the pre-trained detector to process them effectively. Conversely, the output-side normality adaptation module transforms the reconstruction results to match the test-time normal distribution, compensating for any residual shift. This design preserves a consistent latent space, which is crucial for reliable false positive mining, while allowing flexible adaptation.

\paragraph{Input Normality Adaptation.}
Given a multivariate input window $\bm{X}_{t-L+1:t} \in \mathbb{R}^{D \times L}$, we first model the temporal dynamics of each variable independently. For the $i$-th variable, the univariate sequence $\bm{X}_{t-L+1:t}^{(i)} \in \mathbb{R}^L$ is encoded via a 1D convolution layer:
\begin{equation}
\bm{h}_t^{(i)} = \mathrm{Conv}^{\text{in}}(\bm{X}_{t-L+1:t}^{(i)}).
\end{equation}
The resulting temporal embeddings $\{\bm{h}_t^{(i)}\}_{i=1}^{D}$ are then processed by a single inter-variable attention layer to capture cross-variable dependencies:
\begin{equation}
\{\bm{h}_t^{\prime\,(i)}\}_{i=1}^{D} = \mathrm{ATTN}^{\text{in}}(\{\bm{h}_t^{(i)}\}_{i=1}^{D}).
\end{equation}
We apply a variable-wise linear layer to the attention output to compute the adjustment term $\bm{A}_t^{(i)} \in \mathbb{R}^L$ for each variable:
\begin{equation}
\bm{A}_t^{(i)} = \mathrm{Linear}^{\text{in}}(\bm{h}_t^{\prime\,(i)}).
\end{equation}
Finally, a learnable gating parameter $g^{(i)}$, activated by a $\tanh$ function, is used to modulate the adjustment applied to each variable:
\begin{equation}
\tilde{\bm{X}}_{t-L+1:t}^{(i)} = \bm{X}_{t-L+1:t}^{(i)} + \tanh(g^{(i)}) \cdot \bm{A}_t^{(i)}.
\end{equation}
The adapted input $\tilde{\bm{X}}_{t-L+1:t}$ is then passed to the frozen pre-trained anomaly detector $f_\theta$ to obtain the reconstruction:
\begin{equation}
\hat{\bm{X}}_{t-L+1:t} = f_\theta(\tilde{\bm{X}}_{t-L+1:t}).
\end{equation}

\paragraph{Output Normality Adaptation.}
To account for distributional shifts in the reconstructed space, we apply an output normality adaptation module with the same architectural structure. Each variable-wise reconstructed sequence $\hat{\bm{X}}_{t-L+1:t}^{(i)} \in \mathbb{R}^L$ is encoded using a 1D convolution layer:
\begin{equation}
\bm{h}_t^{(i,\text{out})} = \mathrm{Conv}^{\text{out}}(\hat{\bm{X}}_{t-L+1:t}^{(i)}).
\end{equation}
The resulting temporal embeddings $\{\bm{h}_t^{(i,\text{out})}\}_{i=1}^{D}$ are passed through an inter-variable attention layer:
\begin{equation}
\{\bm{h}_t^{\prime\,(i,\text{out})}\}_{i=1}^{D} = \mathrm{ATTN}^{\text{out}}(\{\bm{h}_t^{(i,\text{out})}\}_{i=1}^{D}).
\end{equation}
Each attention output is then passed through a variable-wise linear layer to obtain the adjustment term:
\begin{equation}
\bm{A}_t^{(i,\text{out})} = \mathrm{Linear}^{\text{out}}(\bm{h}_t^{\prime\,(i,\text{out})}).
\end{equation}
Finally, a learnable gating parameter $g^{(i, \text{out})}$ modulates the adjustment via a $\tanh$ activation:
\begin{equation}
\tilde{\hat{\bm{X}}}_{t-L+1:t}^{(i)} = \hat{\bm{X}}_{t-L+1:t}^{(i)} + \tanh(g^{(i, \text{out})}) \cdot \bm{A}_t^{(i,\text{out})}.
\end{equation}

\paragraph{Adaptation Objective.}
Only the parameters of the SANA modules are updated during test-time, while the pre-trained backbone parameters $\bm{\theta}$ remain frozen. We minimize the reconstruction loss over the selected adaptation set $\bm{\mathcal{A}}$:
\begin{equation}
\mathcal{L}_{\text{adapt}} = \frac{1}{D \cdot L} \sum_{\bm{X}_t \in \bm{\mathcal{A}}} \left\| \bm{X}_{t-L+1:t} - \tilde{\hat{\bm{X}}}_{t-L+1:t} \right\|_2^2.
\end{equation}

This modular design enables the model to adapt to both temporal and relational distributional shifts at test-time, while preserving the generalization capabilities of the frozen backbone. By updating only lightweight modules with selectively chosen, informative test samples, our framework achieves robust and efficient test-time adaptation for anomaly detection.

\begin{table*}[t]
\footnotesize
\centering
\begin{tabular}{c c | ccc | ccc | ccc | ccc | ccc}
\toprule
\multirow{2}{*}{\textbf{FPM}} & \multirow{2}{*}{\textbf{SANA}} 
& \multicolumn{3}{c|}{SWaT} 
& \multicolumn{3}{c|}{SMD\_1-8} 
& \multicolumn{3}{c|}{SMD\_2-1} 
& \multicolumn{3}{c}{SMD\_2-4}
& \multicolumn{3}{c}{SMD\_3-2} \\
& & ROC & PRC & F1 & ROC & PRC & F1 & ROC & PRC & F1 & ROC & PRC & F1 & ROC & PRC & F1 \\
\midrule
\multicolumn{2}{c|}{w/o TTA} & 0.83 & 0.72 & 0.29 & 0.72 & 0.33 & 0.10 & 0.65 & 0.28 & 0.27 & 0.82 & 0.46 & 0.32 & 0.45 & 0.16 & 0.25 \\
\midrule
\xmark & \xmark & 0.89 & 0.77 & 0.64 & 0.77 & 0.35 & 0.12 & 0.68 & 0.30 & 0.31 & 0.83 & 0.46 & 0.31 & 0.64 & 0.19 & 0.19 \\
\cmark & \xmark & 0.80 & 0.71 & 0.22 & 0.69 & 0.32 & 0.01 & 0.71 & 0.34 & 0.20 & \textbf{0.93} & \textbf{0.66} & 0.25 & \textbf{0.74} & \textbf{0.22} & 0.21 \\
\xmark & \cmark & 0.89 & 0.78 & 0.64 & 0.83 & 0.36 & 0.15 & 0.71 & 0.32 & 0.33 & 0.83 & 0.46 & 0.37 & 0.68 & 0.19 & 0.25 \\
\cmark & \cmark & 0.89 & 0.78 & 0.62 & \textbf{0.87} & \textbf{0.42} & \textbf{0.21} & \textbf{0.78} & \textbf{0.35} & 0.33 & 0.90 & 0.60 & \textbf{0.51} & 0.72 & 0.20 & \textbf{0.26} \\
\bottomrule
\end{tabular}
\caption{
Ablation study of CANDI across five datasets at $\alpha = 5.0\%$. 
\textbf{FPM} and \textbf{SANA} denote False Positive Mining and Spatiotemporally-Aware Normality Adaptation, respectively.
ROC and PRC denote AUROC and AUPRC, respectively. 
}
\label{tab:ablation}
\vspace{-5pt}
\end{table*}

\section{Experiments}
\subsection{Experimental Setup} 
\paragraph{Datasets.}
We conduct experiments on representative multivariate time-series anomaly detection benchmarks: SWaT~\cite{SWaT} and SMD~\cite{SMD}.
SWaT is industrial control system datasets containing labeled normal and attack periods, reflecting real-world operational and environmental shifts. SMD is a dataset collected from server machines, organized into multiple subdatasets based on the entity.
Since distribution shifts are not uniformly present across all SMD subsets, we select a representative subset of 5 server entities that exhibit prominent normality shifts:
SMD\_1-7, SMD\_1-8, SMD\_2-1, SMD\_2-4 and SMD\_3-2.
\cam{We also evaluate CANDI on the 200 multivariate time-series datasets provided by the TSB-AD benchmark~\cite{elephant2024} to further assess robustness under a broader and more diverse collection of real-world conditions.}

\paragraph{Baselines.}
To assess the benefits of test-time adaptation, we compare our method to M2N2~\cite{kim2024model}, a recent approach that applies TTA to time-series anomaly detection. Following the original setup in M2N2, we use an MLP-based autoencoder as the pre-trained anomaly detector for both methods.
For each dataset, we report the performance of:
(1) the pre-trained model without adaptation,
(2) the model adapted with M2N2, and
(3) the model adapted with our proposed CANDI framework.
This comparison allows us to isolate the effects of different adaptation strategies under the same model backbone.

\paragraph{Implementation Details.}
We evaluate detection performance using the standard metrics: AUROC and AUPRC. In addition, we report F1 scores at fixed false positive rate (FPR) thresholds determined by the $\alpha$-percentile of validation anomaly scores. Specifically, we report results for $\alpha \in \{0.5\%, 1\%, 5\%\}$.
The pre-trained models are trained using the Adam optimizer~\cite{kingma2014adam} with an initial learning rate of 0.001 and cosine learning rate scheduling~\cite{cosineschedule} for 30 epochs.
M2N2 is re-implemented to match our experimental setup for fair comparison. All experiments are conducted using three different random seeds, and we report the average performance. Full results, including standard deviations and additional implementation details, are included in the Appendix.

\subsection{Evaluating Test-time Adaptation in Multivariate Time-series Anomaly Detection}

Table~\ref{tab:main-results} compares the performance of the pre-trained anomaly detector, M2N2, and CANDI under test-time distribution shift on multiple multivariate time-series anomaly detection benchmarks. We report AUROC, AUPRC, and F1 scores across three anomaly score thresholds, $\alpha \in \{0.5\%, 1.0\%, 5.0\%\}$, with $\tau$ set as the $\alpha$-percentile of validation scores. Smaller $\alpha$ yields stricter thresholds with fewer false positives, while larger $\alpha$ reflects more relaxed thresholds, admitting more ambiguous cases.

CANDI consistently matches or outperforms both baselines, with the largest gains seen where the pre-trained model struggles. For instance, on SMD\_1-8, CANDI improves AUROC from 0.719 (pre-trained) and 0.772 (M2N2) to 0.867 at $\alpha = 5.0\%$. Similarly, on SMD\_3-2, AUROC rises from 0.451 to 0.717---a 59.0\% relative improvement---showing CANDI’s strength in challenging conditions.

Notably, at $\alpha = 5.0\%$, where many false positives emerge due to a lower threshold, CANDI turns this challenge into an advantage. The false positive mining selects informative samples close to trusted normal patterns in latent space and adapts using the lightweight SANA module. As a result, CANDI achieves substantial gains; for example, on SMD\_2-4, F1 improves from 0.316 (pre-trained) and 0.311 (M2N2) to 0.512, with AUPRC increasing to 0.600. 
\cam{We provide further evaluation results on the large-scale TSB-AD benchmark~\cite{elephant2024} in the Appendix.}

\subsection{Ablation Study}
Table~\ref{tab:ablation} presents an ablation study evaluating the contributions of the two core components of CANDI: False Positive Mining (FPM) and Spatiotemporally-Aware Normality Adaptation (SANA), across five datasets at $\alpha = 5.0\%$. When FPM is disabled, the model performs adaptation using all test samples whose anomaly scores fall below the threshold. When SANA is disabled, entire parameters of the pre-trained anomaly detector are updated during adaptation. When both FPM and SANA are disabled, the model corresponds to M2N2, which adapts the entire pre-trained anomaly detector using all test samples whose anomaly scores fall below the threshold. This setting serves as a baseline for assessing the effectiveness of each component. 

When FPM is used without SANA, we observe performance degradation across several datasets. For example, on SWaT, F1 drops from 0.64 (M2N2) to 0.22, and on SMD\_2-1, from 0.31 to 0.20. 
This suggests that although FPM improves the adaptation sample quality by mining informative candidates, the adaptation process without the structural constraint of SANA updates distorts the pre-trained latent space, undermining the reliability of FPM's latent similarity calculations. 
These results underscore the necessity of freezing the pre-trained model and adapting only the SANA module to preserve latent consistency.

In contrast, using only SANA without FPM leads to consistent gains over M2N2. 
For instance, on SMD\_2-4, F1 improves from 0.31 to 0.37, and on SMD\_1-8, from 0.12 to 0.15. 
This shows that even without selective sample mining, restricting updates to a lightweight, structurally-informed module like SANA prevents catastrophic forgetting and enables stable test-time adaptation. However, this configuration does not leverage the informative signals present in potential false positives, which limits its ability to fully recover useful patterns missed during training.

The proposed CANDI framework, with both FPM and SANA enabled, achieves the best performance across the majority of datasets despite using fewer adaptation samples than M2N2. 
These results demonstrate that combining sample selection through FPM with the localized and structured updates enabled by SANA provides a robust and efficient adaptation strategy. 
By leveraging latent-consistent false positives while preserving the integrity of the pre-trained detector, CANDI achieves both superior accuracy and stable performance under distribution shift.

\begin{table*}[t]
\centering
\footnotesize
\begin{tabular}{l|l|cc|cc|ccc|ccc}
\toprule
Dataset & Method 
& \begin{tabular}[c]{@{}c@{}}Mod.\\ Samples\end{tabular}
& \begin{tabular}[c]{@{}c@{}}Ano.\\ in Mod.\end{tabular}
& \begin{tabular}[c]{@{}c@{}}Hard\\ Samples\end{tabular}
& \begin{tabular}[c]{@{}c@{}}Ano.\\ in Hard\end{tabular}
& \begin{tabular}[c]{@{}c@{}}Total\\ Adapt\end{tabular}
& \begin{tabular}[c]{@{}c@{}}Total\\ Test\end{tabular}
& \begin{tabular}[c]{@{}c@{}}Total\\ Ano.\end{tabular}
& ROC & PRC & F1 \\
\midrule
\multirow{4}{*}{SMD\_1-8} 
& M2N2                    & N/A     & N/A   & N/A    & N/A  & 11,760 & \multirow{4}{*}{23,690} & \multirow{4}{*}{943} & 0.772 & 0.354 & 0.115 \\
& CANDI Hard       & N/A     & N/A   & 5,437  & 527  & 5,437  &                         & & 0.861 & 0.416 & 0.166 \\
& CANDI Mod.       & 11,038  & 114   & N/A    & N/A  & 11,038 &                         & & 0.826 & 0.362 & 0.142 \\
& CANDI (Hard + Mod.)     & 14,924  & 178   & 4,332  & 497  & 19,256 &                         & & \textbf{0.867} & \textbf{0.423} & \textbf{0.213} \\
\midrule
\multirow{4}{*}{SMD\_2-1} 
& M2N2                    & N/A     & N/A   & N/A    & N/A  & 22,573 & \multirow{4}{*}{23,685} & \multirow{4}{*}{1,287} & 0.683 & 0.296 & 0.309 \\
& CANDI Hard       & N/A     & N/A   & 353    & 78   & 353    &                          & & 0.735 & 0.323 & 0.262 \\
& CANDI Mod.       & 13,940  & 184   & N/A    & N/A  & 13,940 &                          & & 0.717 & 0.317 & 0.328 \\
& CANDI (Hard + Mod.)     & 10,948  & 107   & 310    & 76   & 11,258 &                          & & \textbf{0.780} & \textbf{0.348} & \textbf{0.327} \\
\bottomrule
\end{tabular}
\caption{
Comparison of different adaptation sample configurations on SMD\_1-8 and SMD\_2-1. ``Mod.'' and ``Hard'' refer to the difficulty levels of the samples used. ``Ano.'' refers to anomaly. ``Total Adapt'' denotes the number of adaptation samples.
}
\label{tab:adaptation-sample-config}
\end{table*}

\begin{figure}[t]
    \centering
    \begin{minipage}[b]{0.49\columnwidth}
        \centering
        \includegraphics[width=\linewidth]{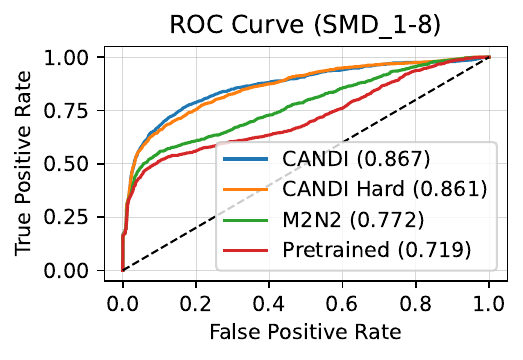}
        \label{fig:roc_1}
    \end{minipage}
    \begin{minipage}[b]{0.49\columnwidth}
        \centering
        \includegraphics[width=\linewidth]{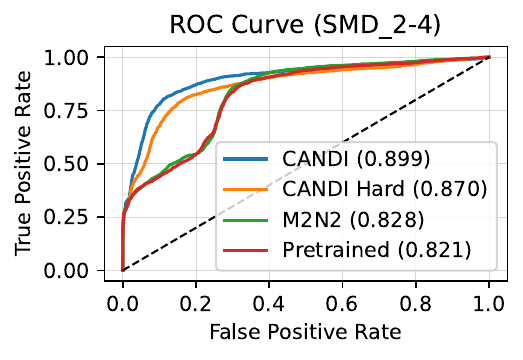}
        \label{fig:roc_2}
    \end{minipage}
    
    \vspace{-0.75em}

    \begin{minipage}[b]{0.49\columnwidth}
        \centering
        \includegraphics[width=\linewidth]{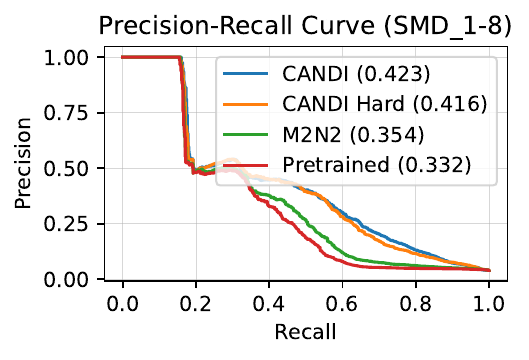}
        \label{fig:pr_1}
    \end{minipage}
    \begin{minipage}[b]{0.49\columnwidth}
        \centering
        \includegraphics[width=\linewidth]{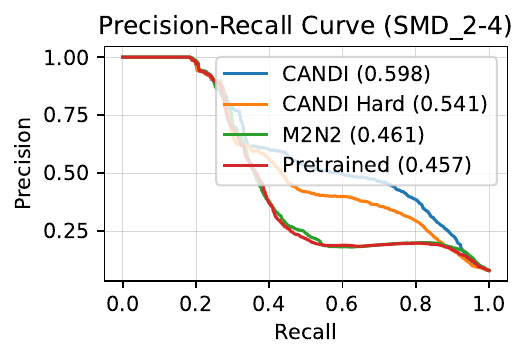}
        \label{fig:pr_2}
    \end{minipage}
    \vspace{-15pt}
    \caption{Receiver Operating Characteristic (ROC) and Precision-Recall (PR) curves for anomaly detection. The value in parentheses indicates the area under each curve.}
    \vspace{-10pt}
    \label{fig:roc_pr}
\end{figure}

\paragraph{Analysis on ROC and PR Curves.}
Figure~\ref{fig:roc_pr} shows the Receiver Operating Characteristic (ROC) and Precision-Recall (PR) curves to further compare the detection capabilities. ROC curves show that CANDI consistently achieves a significantly higher true positive rate (TPR) at low false positive rates (\textit{e.g.}, FPR = 0.1) compared to all baselines. This indicates that CANDI is more effective at identifying true anomalies while keeping false alarms low—a critical property in real-world deployment scenarios.

We also evaluate CANDI Hard, a variant that uses only the hard samples identified through false positive mining, excluding moderate samples. Despite using fewer adaptation samples, this variant still outperforms the baselines and achieves performance close to the original CANDI. These results highlight that even partial adaptation guided by carefully curated samples can offer substantial benefits, and that the CANDI framework further enhances performance by incorporating additional reliable test-time samples.

\paragraph{Comparison of Adaptation Sample Configurations.}
Table~\ref{tab:adaptation-sample-config} compares the effectiveness of different adaptation sample configurations, including the proposed CANDI variants and M2N2. Notably, CANDI with only hard samples outperforms M2N2 while using significantly fewer adaptation samples---less than half on SMD\_1-8 (5,437 vs. 11,760) and less than 2\% on SMD\_2-1 (353 vs. 22,573). Despite this drastic reduction, it achieves superior AUROC and AUPRC, highlighting the efficiency of our curated adaptation.

Among the samples identified as potential false positives for adaptation, some fraction are actual anomalies: approximately 10\% on SMD\_1-8 and 25\% on SMD\_2-1. This indicates that CANDI sometimes performs adaptation on mislabeled anomalous data. Nevertheless, it still outperforms the baseline, suggesting robustness to moderate contamination. 
These results highlight that enhancing the accuracy of false positive mining, thereby reducing the potential negative impact of anomaly adaptation, is an important future direction for TTA frameworks in MTSAD.

Among all configurations, the best performance is achieved when both moderate and hard samples are used together. The results demonstrate that combining reliable moderate samples with carefully mined hard samples enables more comprehensive and effective test-time adaptation under distribution shift.

\begin{table}[t]
\centering
\footnotesize
\begin{tabular}{l|ccc|ccc}
\toprule
\multirow{2}{*}{Method} 
& \multicolumn{3}{c|}{SMD\_1-8} 
& \multicolumn{3}{c}{SMD\_2-1} \\
& ROC & PRC & F1 & ROC & PRC & F1 \\
\midrule
Linear & 0.840 & 0.415 & 0.115 & 0.673 & 0.294 & 0.275 \\
SANA   & 0.867 & 0.423 & 0.213 & 0.780 & 0.348 & 0.327 \\
\bottomrule
\end{tabular}
\caption{
Comparison between Linear and SANA adaptation modules on SMD\_1-8 and SMD\_2-1. ROC and PRC denote AUROC and AUPRC, respectively.
}
\vspace{-10pt}
\label{tab:sana-vs-linear}
\end{table}

\paragraph{Effectiveness of SANA Architecture.}
Table~\ref{tab:sana-vs-linear} compares the proposed SANA module with a linear adaptation head on SMD\_1-8 and SMD\_2-1. Across all metrics, SANA outperforms the linear approach—achieving notably higher F1 scores (0.213 vs. 0.115 on SMD\_1-8 and 0.327 vs. 0.275 on SMD\_2-1). This indicates that SANA provides more effective test-time adaptation.
The performance gap highlights the importance of structure-aware adaptation. Unlike linear updates, SANA captures temporal and variable-wise dependencies while preserving the pre-trained model’s latent space. This allows SANA to adapt meaningfully under distribution shifts without degrading the original detector.

\section{Conclusion}
Multivariate time-series anomaly detection in deployment environments suffers from performance degradation due to distribution shift. To address this, we proposed \textbf{CANDI}, a test-time adaptation framework that curates informative false positives and adapts using a lightweight, structure-aware module.
CANDI combines \textit{False Positive Mining} (FPM) to identify reliable adaptation samples with \textit{Spatiotemporally-Aware Normality Adaptation} (SANA), a plug-and-play module that preserves pre-trained knowledge. Experiments show that CANDI significantly outperforms prior methods, especially under relaxed anomaly thresholds.

\section{Acknowledgments}
This work was supported by Institute of Information \& communications Technology Planning \& Evaluation (IITP) grant funded by the Korea government (MSIT) [No.RS-2021-II211343, Artificial Intelligence Graduate School Program (Seoul National University); No.RS-2024-00357879, AI-based Biosignal Fusion and Generation Technology for Intelligent Personalized Chronic Disease Management],
the National Research Foundation of Korea (NRF) grant funded by the Korea government (MSIT)
(No. 2022R1A3B1077720; 2022R1A5A708390811; No.RS-2023-00212484, xAI for Motion Prediction in Complex, Real-World Driving Environment), the BK21 FOUR program of the Education
and the Research Program for Future ICT Pioneers, Seoul National University in 2025, Hyundai Motor Company, and Samsung Electronics Co., Ltd (IO250624-13143-01).

\bibliography{aaai2026}

\clearpage
\onecolumn

\begin{center}
  {\Large \textbf{Supplementary Material for CANDI}}
\end{center}

\section{Dataset Characteristics}
We evaluate our method on widely adopted benchmarks for multivariate time-series anomaly detection (MTSAD): the SWaT~\cite{SWaT} and SMD~\cite{SMD} datasets. These datasets are curated from distinct domains—industrial control and IT infrastructure—and offer realistic, labeled time-series data that reflect complex system dynamics. Their domain-specific characteristics and high-fidelity anomaly labels make them essential for developing and benchmarking robust anomaly detection algorithms.

The SWaT dataset, created by the iTrust research group, captures the behavior of a water treatment plant operating under an industrial control system (ICS). It comprises sensor and actuator readings recorded during both normal operation and simulated cyber-attacks. These attacks are deliberately designed to induce faults in the system, providing ground truth labels for abnormal events. The intricate dependencies among variables and the presence of subtle anomalies make SWaT a challenging and widely used benchmark in ICS anomaly detection research.

On the other hand, the Server Machine Dataset (SMD) reflects monitoring data from large-scale server clusters. It contains multivariate metrics such as CPU usage, memory load, and disk activity, collected over long time periods. The dataset includes both normal and anomalous intervals, with anomalies representing issues like hardware failures or performance bottlenecks. SMD is well-suited for evaluating anomaly detection in IT operations and has become a standard benchmark for studying distribution shifts and temporal variability in system behavior.

\begin{table*}[ht]
\centering
\begin{tabular}{ccccccc}
\toprule
Dataset   & Variables & Train Steps & Test Steps & Anomaly Ratio & ADF Statistics & $p$-value \\
\midrule
SWaT      & 51  & 495,000  & 449,919  & 0.121  & -0.780  & 0.83  \\
SMD\_1-7      & 38  & 23,697 & 23,697  & 0.101  & -2.474  & 0.12  \\
SMD\_1-8      & 38  & 23,698  & 23,699  & 0.032  & -3.305  & 0.01  \\
SMD\_2-1  & 38  & 23,693   & 23,694   & 0.049  & -3.211 & 0.02  \\
SMD\_2-4  & 38  & 23,689   & 23,689   & 0.072  & -3.650 & 0.00  \\
SMD\_3-2  & 38  & 23,702   & 23,703   & 0.047  & -1.430 & 0.56  \\
\bottomrule
\end{tabular}
\caption{Summary of the statistical characteristics of MTSAD datasets: the number of variables, the total time steps of train and test data, the ratio of anomalous time steps in test data, and the degree of stationarity measured using Augmented Dickey-Fuller (ADF) test statistics and the corresponding $p$-value.}
\label{tab:data}
\end{table*}

\begin{figure*}[ht]
    \centering
    \begin{minipage}[b]{0.49\textwidth}
        \centering
        \includegraphics[width=\linewidth]{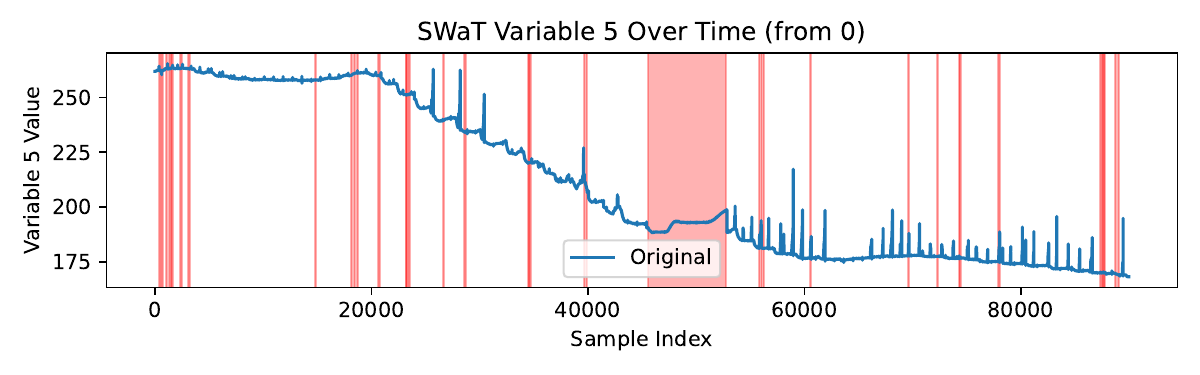}
        \label{fig:data_swat}
    \end{minipage}
    \begin{minipage}[b]{0.49\textwidth}
        \centering
        \includegraphics[width=\linewidth]{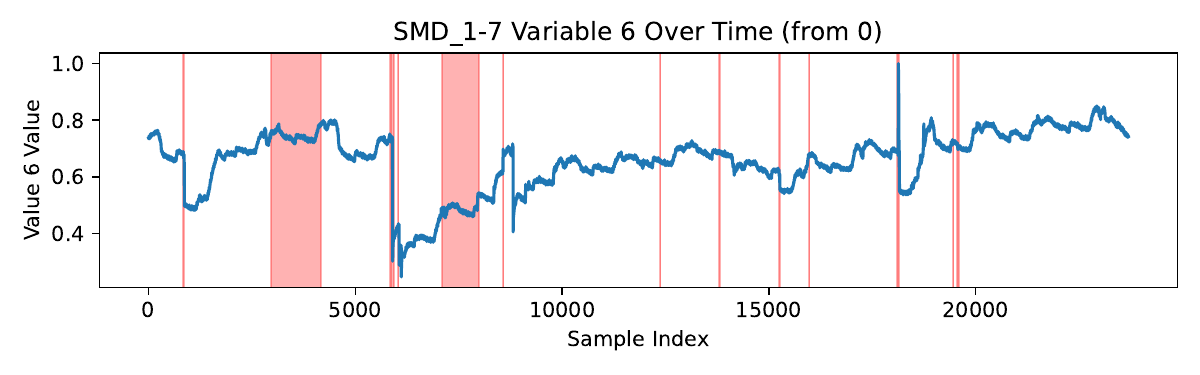}
        \label{fig:data_smd_1_7}
    \end{minipage}
    
    \vspace{-0.75em}

    \begin{minipage}[b]{0.49\textwidth}
        \centering
        \includegraphics[width=\linewidth]{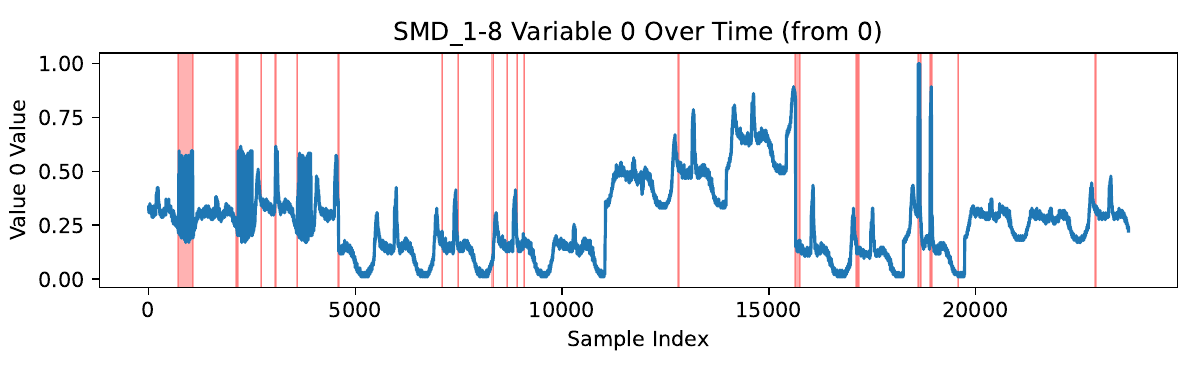}
        \label{fig:data_smd_1_8}
    \end{minipage}
    \begin{minipage}[b]{0.49\textwidth}
        \centering
        \includegraphics[width=\linewidth]{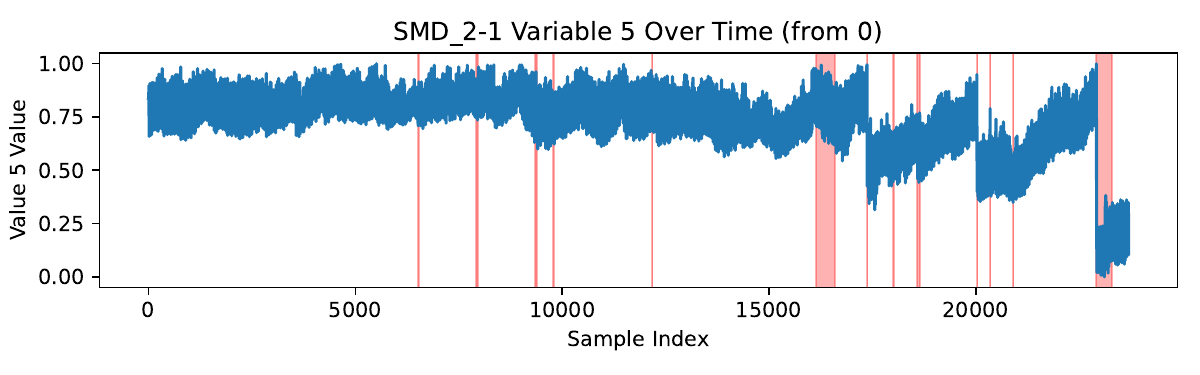}
        \label{fig:data_smd_2_1}
    \end{minipage}

    \vspace{-0.75em}

    \begin{minipage}[b]{0.49\textwidth}
        \centering
        \includegraphics[width=\linewidth]{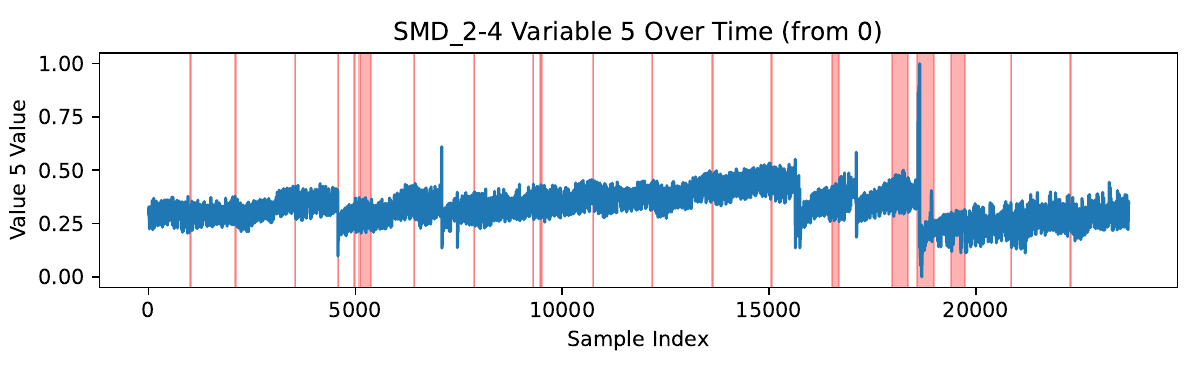}
        \label{fig:data_smd_2_4}
    \end{minipage}
    \begin{minipage}[b]{0.49\textwidth}
        \centering
        \includegraphics[width=\linewidth]{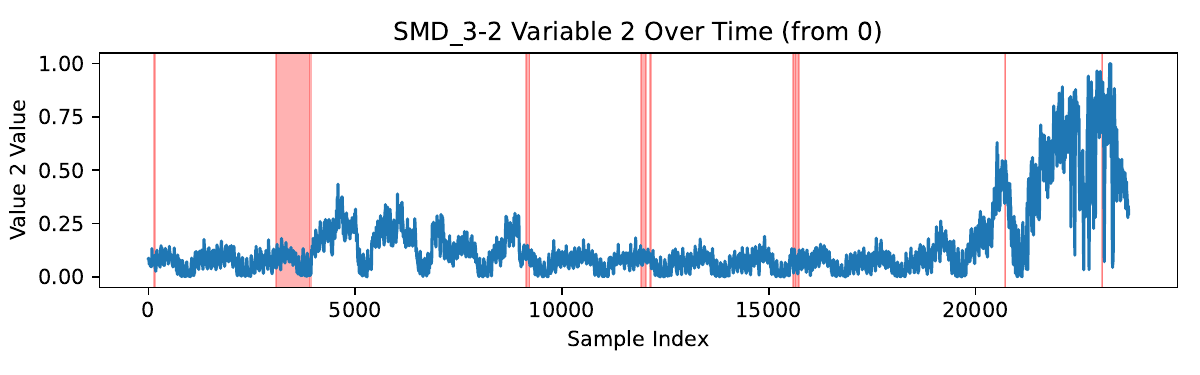}
        \label{fig:data_smd_3_2}
    \end{minipage}

    \caption{Examples of distribution shifts in test data, with red shaded areas indicating true anomalies.}
    \label{fig:data_example}
\end{figure*}

Table~\ref{tab:data} summarizes the key characteristics of the MTSAD datasets used in our study, including the number of variables, the number of time steps for training and testing, the proportion of anomalies in the test set, and the stationarity of the data assessed using the Augmented Dickey-Fuller (ADF) test. The ADF test evaluates whether a time-series is stationary by testing for the presence of a unit root; a low $p$-value (typically $< 0.05$) indicates that the time-series is likely stationary, while a high $p$-value suggests non-stationarity. 
As shown, the SWaT dataset exhibits strong non-stationarity (ADF statistic = $-0.780$, $p$-value = $0.83$), while datasets like SMD\_2-4 show stronger stationarity (ADF statistic = $-3.650$, $p$-value = $0.00$). This variation in stationarity across datasets may influence the behavior and performance of time-series anomaly detection models.

For each dataset, we perform the ADF test on every variable individually and report the ADF statistic and the corresponding $p$-value of the variable with the largest $p$-value, representing the least stationary dimension in the dataset. This conservative reporting approach helps highlight the worst-case stationarity scenario within each dataset, which may be critical for model robustness. As shown, the SWaT dataset exhibits strong non-stationarity (ADF statistic = $-0.780$, $p$-value = $0.83$), while datasets like SMD\_2-4 show stronger stationarity (ADF statistic = $-3.650$, $p$-value = $0.00$). This variation in stationarity across datasets may influence the behavior and performance of time-series anomaly detection models. Figure~\ref{fig:data_example} illustrates examples of distribution shifts observed in the test data of each dataset. The red shaded regions indicate periods where anomalies actually occurred.

\section{Additional Implementation Details}
In all experiments, we set the window length to $L = 10$ and used a batch size of 256 for both training and testing. Since none of the datasets provided an explicit validation split, we used the last 20\% of the training data as validation data. 
For the SWaT dataset, due to its large size, we applied a downsampling rate of 5. To ensure stability during test-time adaptation, we employed gradient clipping with a norm threshold of 0.5. Following common practices in the test-time adaptation literature, we did not apply a single-step update but instead introduced the number of adaptation steps as a hyperparameter.
Adaptation was performed using stochastic gradient descent (SGD) with Nesterov momentum of 0.9 and a weight decay of 0.0001. For CANDI, we used a hidden dimension of $h = 512$ for the SANA module and conducted hyperparameter tuning over learning rates of 0.001, 0.003, 0.01, 0.03, and 0.1; gating parameter initialization values of 0.0, 0.1, and 0.5; and adaptation steps of 1 and 5. For M2N2, we searched over the same set of learning rates and evaluated exponential moving average (EMA) decay rates $\gamma$ of 0.9, 0.99, 0.999, and 0.9999 for the EMA update.
\cam{In CANDI, both hard and moderate samples were accumulated into their respective curated adaptation sets, and adaptation was performed only when at least 16 samples had been gathered, ensuring stable model updates. To further support reliable adaptation, all latent representations were L2-normalized. This promotes a Gaussian-like latent distribution, as shown in Mahalanobis++~\cite{Mahalanobis++}, and stabilizes the distance-based false-positive selection used in our thresholding strategy.}
All experiments were conducted on a single NVIDIA A40 GPU.

\section{Additional Experimental Results}

\subsection{Parameter-Efficient Adaptation via Lightweight SANA}
\begin{table*}[ht]
\centering
\begin{tabular}{cccccccc}
\toprule
\multirow{2}{*}{$h$ Dimension} & \multirow{2}{*}{Additional Params (MB)} & \multicolumn{3}{c}{SMD\_1-8} & \multicolumn{3}{c}{SMD\_2-4} \\
\cmidrule(lr){3-5} \cmidrule(lr){6-8}
 & & AUROC & AUPRC & F1 & AUROC & AUPRC & F1 \\
\midrule
Pre-trained & 0.00 & 0.719 & 0.332 & 0.092 & 0.821 & 0.457 & 0.316 \\
\midrule
512 & 7.07 & 0.867 & 0.423 & 0.213 & 0.899 & 0.600 & 0.512 \\
256 & 2.03 & 0.878 & 0.423 & 0.195 & 0.883 & 0.564 & 0.426 \\
128 & 0.64 & 0.853 & 0.400 & 0.156 & 0.861 & 0.537 & 0.348 \\
64  & 0.23 & 0.851 & 0.389 & 0.129 & 0.849 & 0.522 & 0.312 \\
32  & 0.09 & 0.825 & 0.375 & 0.109 & 0.831 & 0.480 & 0.308 \\
\bottomrule
\end{tabular}
\caption{Performance comparison across different hidden dimensions $h$. As $h$ decreases, the number of additional parameters is reduced, with a moderate trade-off in performance for SMD\_1-8 and SMD\_2-4. The first row shows the performance of the frozen pre-trained anomaly detector without TTA.}
\label{tab:hidden_dim}
\end{table*}

Table~\ref{tab:hidden_dim} presents the performance of CANDI with varying hidden dimensions $h$ in the SANA module, which directly determines the number of additional parameters used for test-time adaptation. The results show that CANDI achieves substantial performance gains over the pre-trained model even with minimal parameter overhead. For example, with just 2.03MB of additional parameters ($h=256$), CANDI improves AUROC from 0.719 to 0.878 on SMD\_1-8, and from 0.821 to 0.883 on SMD\_2-4, surpassing the pre-trained model by up to +22.1\% and +7.6\%, respectively. Notably, even with a very small footprint of only 0.64MB ($h=128$), CANDI still improves AUROC by +13.4\% and +4.0\% over the non-adaptive baseline.

As $h$ decreases further, the number of additional parameters drops significantly (e.g., 0.09MB at $h=32$), and although performance degrades slightly, the results still outperform the pre-trained model in most metrics. For instance, at $h=32$, CANDI's F1 on SMD\_1-8 is 0.109, which is +18.5\% higher than the pre-trained baseline (0.092), despite requiring less than 0.1MB of additional parameters.

These results confirm that CANDI enables highly parameter-efficient test-time adaptation. With minimal increases in model size, it significantly boosts anomaly detection accuracy across multiple benchmarks, making it practical for deployment in resource-constrained settings such as embedded systems or edge devices.

\subsection{Stability Across Random Seeds}

\begin{table*}[ht]
\footnotesize
\centering
\begin{tabular}{l|l|ccc|ccc|ccc}
\toprule
\multirow{2}{*}{Dataset} & \multirow{2}{*}{Metric (Std)} & \multicolumn{3}{c|}{$\alpha = 0.5\%$} & \multicolumn{3}{c|}{$\alpha = 1.0\%$} & \multicolumn{3}{c}{$\alpha = 5.0\%$} \\
& & Pre. & M2N2 & CANDI (p) & Pre. & M2N2 & CANDI (p) & Pre. & M2N2 & CANDI (p) \\
\midrule
\multirow{3}{*}{SWaT}
& AUROC  & 0.000 & 0.000 & 0.005 (0.35) & 0.000 & 0.000 & 0.002 (0.12) & 0.000 & 0.000 & 0.002 (0.09) \\
& AUPRC  & 0.000 & 0.000 & 0.004 (0.03) & 0.000 & 0.000 & 0.001 (0.00) & 0.000 & 0.000 & 0.002 (0.01) \\
& F1     & 0.000 & 0.000 & 0.002 (0.00) & 0.000 & 0.000 & 0.000 (0.00) & 0.000 & 0.000 & 0.001 (0.00) \\
\midrule
\multirow{3}{*}{SMD\_1-7}
& AUROC  & 0.000 & 0.000 & 0.018 (0.96) & 0.000 & 0.000 & 0.018 (0.07) & 0.000 & 0.000 & 0.018 (0.25) \\
& AUPRC  & 0.000 & 0.000 & 0.019 (0.35) & 0.000 & 0.000 & 0.019 (0.23) & 0.000 & 0.000 & 0.019 (0.81) \\
& F1     & 0.000 & 0.000 & 0.002 (0.00) & 0.000 & 0.000 & 0.115 (0.18) & 0.000 & 0.000 & 0.001 (0.07) \\
\midrule
\multirow{3}{*}{SMD\_1-8}
& AUROC  & 0.000 & 0.000 & 0.001 (0.00) & 0.000 & 0.000 & 0.001 (0.00) & 0.000 & 0.000 & 0.002 (0.00) \\
& AUPRC  & 0.000 & 0.000 & 0.002 (0.00) & 0.000 & 0.000 & 0.002 (0.00) & 0.000 & 0.000 & 0.001 (0.00) \\
& F1     & 0.000 & 0.000 & 0.001 (0.00) & 0.000 & 0.000 & 0.002 (0.00) & 0.000 & 0.000 & 0.006 (0.00) \\
\midrule
\multirow{3}{*}{SMD\_2-1}
& AUROC  & 0.000 & 0.000 & 0.004 (0.01) & 0.000 & 0.000 & 0.002 (0.00) & 0.000 & 0.000 & 0.041 (0.15) \\
& AUPRC  & 0.000 & 0.000 & 0.002 (0.02) & 0.000 & 0.000 & 0.001 (0.00) & 0.000 & 0.000 & 0.021 (0.12) \\
& F1     & 0.000 & 0.000 & 0.000 (0.00) & 0.000 & 0.000 & 0.001 (0.01) & 0.000 & 0.000 & 0.011 (0.09) \\
\midrule
\multirow{3}{*}{SMD\_2-4}
& AUROC  & 0.000 & 0.000 & 0.001 (0.00) & 0.000 & 0.000 & 0.000 (0.00) & 0.000 & 0.000 & 0.009 (0.01) \\
& AUPRC  & 0.000 & 0.000 & 0.001 (0.07) & 0.000 & 0.000 & 0.001 (0.02) & 0.000 & 0.000 & 0.034 (0.03) \\
& F1     & 0.000 & 0.000 & 0.002 (0.42) & 0.000 & 0.000 & 0.009 (0.17) & 0.000 & 0.000 & 0.025 (0.01) \\
\midrule
\multirow{3}{*}{SMD\_3-2}
& AUROC  & 0.000 & 0.000 & 0.006 (0.00) & 0.000 & 0.000 & 0.006 (0.00) & 0.000 & 0.000 & 0.008 (0.01) \\
& AUPRC  & 0.000 & 0.000 & 0.002 (0.00) & 0.000 & 0.000 & 0.002 (0.01) & 0.000 & 0.000 & 0.002 (0.02) \\
& F1     & 0.000 & 0.000 & 0.000 (0.00) & 0.000 & 0.000 & 0.000 (0.00) & 0.000 & 0.000 & 0.000 (0.00) \\
\bottomrule
\end{tabular}
\caption{Standard deviation and p-values of test-time adaptation performance across three random seeds. Pre. denotes a pretrained model without test-time adaptation. Pretrained and M2N2 are deterministic (std = 0.000). p-values indicate the statistical significance of CANDI's performance over M2N2.}
\label{tab:main-std-pval}
\end{table*}

To assess the robustness of CANDI under different initializations, we report the standard deviation of its performance across three random seeds in Table~\ref{tab:main-std-pval}. CANDI exhibits consistently low variance across AUROC, AUPRC, and F1 metrics, with standard deviation values generally below 0.01 across most datasets and threshold settings. This indicates that CANDI maintains stable performance despite the inherent stochasticity of test-time adaptation, ensuring reliable and consistent detection under distribution shift.

To further assess the statistical reliability of these gains, we conducted paired t-tests comparing CANDI and M2N2 results. The table includes p-values computed from these tests. In many cases—particularly on SMD\_1-8, SMD\_2-1, and SMD\_3-2—the improvements by CANDI are statistically significant (e.g., $p < 0.05$), indicating that the observed gains are unlikely to be due to chance. Notably, AUPRC and F1 scores show strong significance on these datasets. However, on certain benchmarks like SMD\_1-7 and SMD\_2-4 under specific threshold settings, the p-values are higher, suggesting that performance improvements are less consistent or marginal in those cases. 
These results validate that CANDI not only performs stably but also provides statistically meaningful improvements over baselines in most scenarios.

\subsection{Hyperparameter Robustness Analysis}
\begin{table}[ht]
\centering
\begin{tabular}{c|c|c|c|c}
\toprule
lr & gating\_init & AUROC & AUPRC & F1 \\
\midrule
\multirow{3}{*}{0.03}
 & 0.5 & 0.871 & 0.424 & 0.223 \\
 & 0.1 & 0.860 & 0.403 & 0.254 \\
 & 0.0 & 0.862 & 0.398 & 0.227 \\
\midrule
\multirow{3}{*}{0.01}
 & 0.5 & 0.878 & 0.411 & 0.177 \\
 & 0.1 & 0.852 & 0.400 & 0.205 \\
 & 0.0 & 0.827 & 0.390 & 0.155 \\
\midrule
\multirow{3}{*}{0.003}
 & 0.5 & 0.872 & 0.418 & 0.221 \\
 & 0.1 & 0.796 & 0.367 & 0.123 \\
 & 0.0 & 0.735 & 0.342 & 0.094 \\
\bottomrule
\end{tabular}
\caption{Robustness of CANDI to learning rate and gating parameter initialization on SMD\_1-8.}
\label{tab:hyperparam-robustness}
\end{table}

\begin{table}[ht]
\centering
\footnotesize
\begin{tabular}{lccc}
\toprule
Batch Size & AUROC & AUPRC & F1 \\
\midrule
64  & 0.91 & 0.72 & 0.73 \\
128 & 0.92 & 0.73 & 0.73 \\
256 & 0.93 & 0.74 & 0.73 \\
512 & 0.92 & 0.74 & 0.73 \\
\bottomrule
\end{tabular}
\caption{Sensitivity of CANDI to different test-time batch sizes on the SMD\_1-7 dataset.}
\label{tab:test-batchsize}
\end{table}

Table~\ref{tab:hyperparam-robustness} shows the performance of CANDI under different combinations of learning rate and gating parameter initialization on SMD\_1-8. We observe that CANDI consistently achieves high performance across a wide range of hyperparameter settings, indicating strong robustness.
In particular, performance remains stable when the learning rate is set to 0.03 or 0.003, with AUROC ranging from 0.860 to 0.872 and AUPRC from 0.398 to 0.424 across different gating initializations. While the F1 score shows some variation depending on the specific combination, CANDI still outperforms the baseline by a large margin in most settings.

\cam{We also evaluate the sensitivity of CANDI to the test-time batch size, an important factor in test-time adaptation since batch size determines the amount of temporal context available and often varies in practice due to latency or memory constraints. As shown in Table~\ref{tab:test-batchsize}, CANDI maintains stable performance across batch sizes ranging from 64 to 512, with AUROC increasing moderately from 0.91 to 0.93 and AUPRC from 0.72 to 0.74, while the F1 score remains essentially unchanged at 0.73. These results demonstrate that CANDI’s adaptation dynamics are not overly dependent on test-time batch size, further confirming its robustness under practical test-time conditions.}

\subsection{Effectiveness of CANDI Across Architectures}

\begin{table*}[ht]
\centering
\begin{tabular}{l|l|ccc|ccc|ccc}
\toprule
\textbf{Model} & \textbf{Metric} & \multicolumn{3}{c|}{SMD\_1-8} & \multicolumn{3}{c|}{SMD\_2-4} & \multicolumn{3}{c}{SMD\_3-2} \\
& & Pretrained & M2N2 & CANDI & Pretrained & M2N2 & CANDI & Pretrained & M2N2 & CANDI \\
\midrule
\multirow{3}{*}{MLP}
& AUROC & 0.719 & 0.772 & 0.867 & 0.821 & 0.828 & 0.899 & 0.451 & 0.640 & 0.717 \\
& AUPRC & 0.332 & 0.354 & 0.423 & 0.457 & 0.461 & 0.600 & 0.159 & 0.188 & 0.199 \\
& F1     & 0.092 & 0.115 & 0.213 & 0.316 & 0.311 & 0.512 & 0.247 & 0.194 & 0.262 \\
\midrule
\multirow{3}{*}{TimesNet}
& AUROC & 0.606 & 0.683 & 0.910 & 0.762 & 0.818 & 0.762 & 0.712 & 0.740 & 0.769 \\
& AUPRC & 0.256 & 0.305 & 0.375 & 0.388 & 0.523 & 0.388 & 0.145 & 0.162 & 0.162 \\
& F1     & 0.075 & 0.112 & 0.153 & 0.180 & 0.299 & 0.180 & 0.065 & 0.054 & 0.106 \\
\bottomrule
\end{tabular}
\caption{Performance comparison of test-time adaptation methods across three SMD benchmarks for different model architectures.}
\label{tab:tta_comparison}
\end{table*}

Table~\ref{tab:tta_comparison} presents a comprehensive comparison of test-time adaptation methods—Pretrained (no adaptation), M2N2, and CANDI—applied to two different backbone architectures (MLP and TimesNet~\cite{wu2022timesnet}) across three subsets of the SMD dataset. 
Across most settings, CANDI consistently outperforms both Pretrained and M2N2 baselines, demonstrating its effectiveness and generalizability. For example, with the MLP model on SMD\_1-8, CANDI improves AUROC from 0.719 (Pretrained) and 0.772 (M2N2) to 0.867. Similarly, on the challenging SMD\_3-2 subset, it shows substantial gains across all metrics, highlighting its robustness under severe distribution shifts. 
For TimesNet, which has strong baseline performance, CANDI still brings noticeable improvements—particularly on SMD\_1-8, where AUROC improves from 0.606 (Pretrained) and 0.683 (M2N2) to 0.910. These gains indicate that even high-performing models can benefit from CANDI's targeted adaptation mechanism. 
However, we observe relatively marginal improvements on SMD\_2-4 with TimesNet, where the performance of CANDI nearly matches that of the Pretrained model. One possible explanation is that SMD\_2-4 exhibits minimal distribution shift between training and test data, reducing the benefit of test-time adaptation. In such cases, the initial model may already generalize well, leaving little room for improvement. Another possibility is that the test-time false positives in SMD\_2-4 are less informative or less concentrated, making it harder for the False Positive Mining module to curate effective adaptation samples. 
Despite this, CANDI shows no sign of performance degradation, demonstrating its safety and stability even when adaptation brings limited gains. These results highlight CANDI's robustness and adaptability across varying degrees of distribution shift and model architectures.

\subsection{\cam{Evaluation on Large-Scale MTSAD Benchmark}}
\cam{To further validate the generality and robustness of CANDI, we additionally evaluated it on the large-scale TSB-AD benchmark~\cite{elephant2024}, which provides over 1000 carefully curated TSAD datasets and addresses several well-known limitations of prior benchmarks. This benchmark covers diverse conditions, including both non-stationary and relatively stationary time-series datasets, enabling a more comprehensive evaluation of model behavior. In this setting, we report not only standard metrics such as AUPRC and AUROC but also VUS-PR and VUS-ROC~\cite{vus2022}, as these metrics offer a more holistic assessment of anomaly-detection performance. Traditional threshold-based metrics like F1 are highly sensitive to the choice of anomaly-score threshold, and even threshold-free metrics such as AUROC and AUPRC remain affected by how strictly anomaly boundaries are defined in time series. VUS-PR and VUS-ROC mitigate this boundary-sensitivity by evaluating PR and ROC curves across a full range of plausible boundary tolerances, producing a continuous surface over thresholds and buffer widths. Following the protocol in~\cite{elephant2024}, and because our method targets multivariate TSAD, we conducted evaluations on the designated subset of 200 multivariate datasets.}

\cam{Across all five evaluation metrics—AUPRC, AUROC, VUS-PR, VUS-ROC, and F1—CANDI achieves the strongest performance among all competing methods, as shown in Table~\ref{tab:elephant_benchmark}. Notably, the improvements on VUS-PR and VUS-ROC are particularly substantial, outperforming both established MTSAD baselines such as OmniAnomaly~\cite{OmniAnomaly}, TranAD~\cite{TranAD}, and TimesNet~\cite{wu2022timesnet}, as well as recent adaptation-based approaches including M2N2~\cite{kim2024model}. These results indicate that CANDI not only excels on conventional MTSAD benchmarks but also sustains top-tier performance under the more rigorous and diverse evaluation setting of~\cite{elephant2024}, underscoring its reliability and scalability for real-world multivariate anomaly detection.}

\begin{table}[ht]
\centering
\footnotesize
\begin{tabular}{lccccc}
\toprule
Model & AUPRC & AUROC & VUS-PR & VUS-ROC & F1 \\
\midrule
CNN           & 0.32 & 0.73 & 0.31 & 0.76 & 0.37 \\
OmniAnomaly   & 0.27 & 0.65 & 0.31 & 0.69 & 0.32 \\
TranAD        & 0.14 & 0.59 & 0.18 & 0.65 & 0.21 \\
TimesNet      & 0.13 & 0.56 & 0.19 & 0.64 & 0.20 \\
M2N2          & 0.32 & 0.74 & 0.32 & 0.78 & 0.38 \\
CANDI         & 0.35 & 0.75 & 0.42 & 0.80 & 0.43 \\
\bottomrule
\end{tabular}
\caption{Performance on the multivariate subset of the TSB-AD benchmark, which comprises 200 carefully curated multivariate time-series datasets. Alongside standard metrics (AUPRC, AUROC, F1), we also report VUS-PR and VUS-ROC, which offer boundary-tolerance–aware, parameter-free evaluation of anomaly detection performance.}
\label{tab:elephant_benchmark}
\end{table}

\subsection{Qualitative Results of CANDI on Anomaly Scoring}
\begin{figure*}[ht]
    \centering
    \begin{minipage}[b]{0.49\textwidth}
        \centering
        \includegraphics[width=\linewidth]{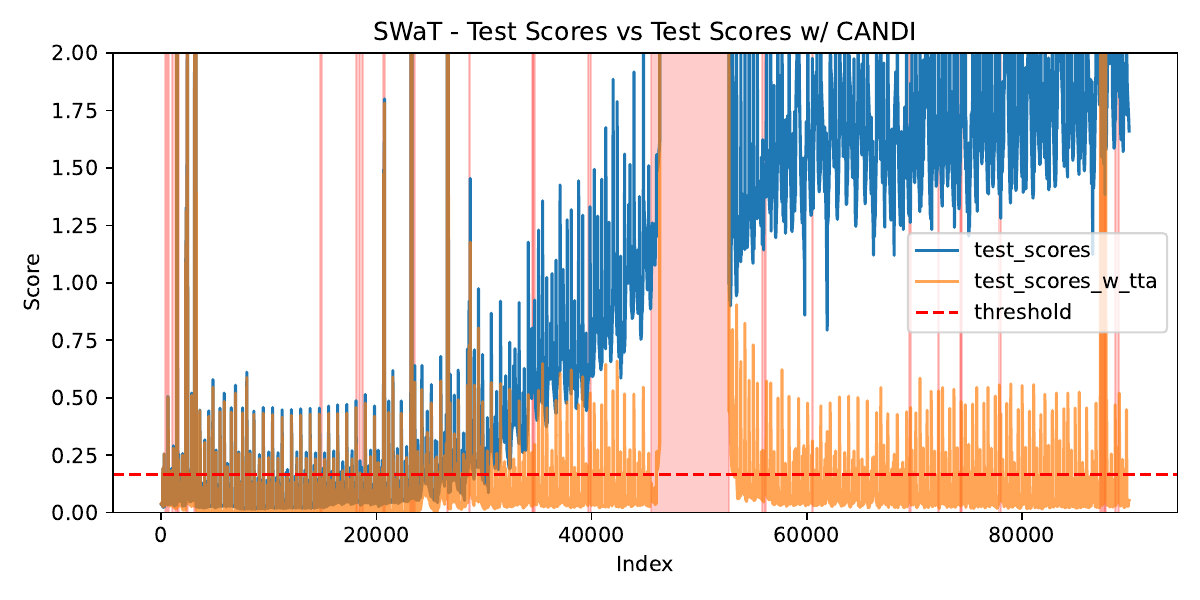}
        \label{fig:scores_swat}
    \end{minipage}
    \begin{minipage}[b]{0.49\textwidth}
        \centering
        \includegraphics[width=\linewidth]{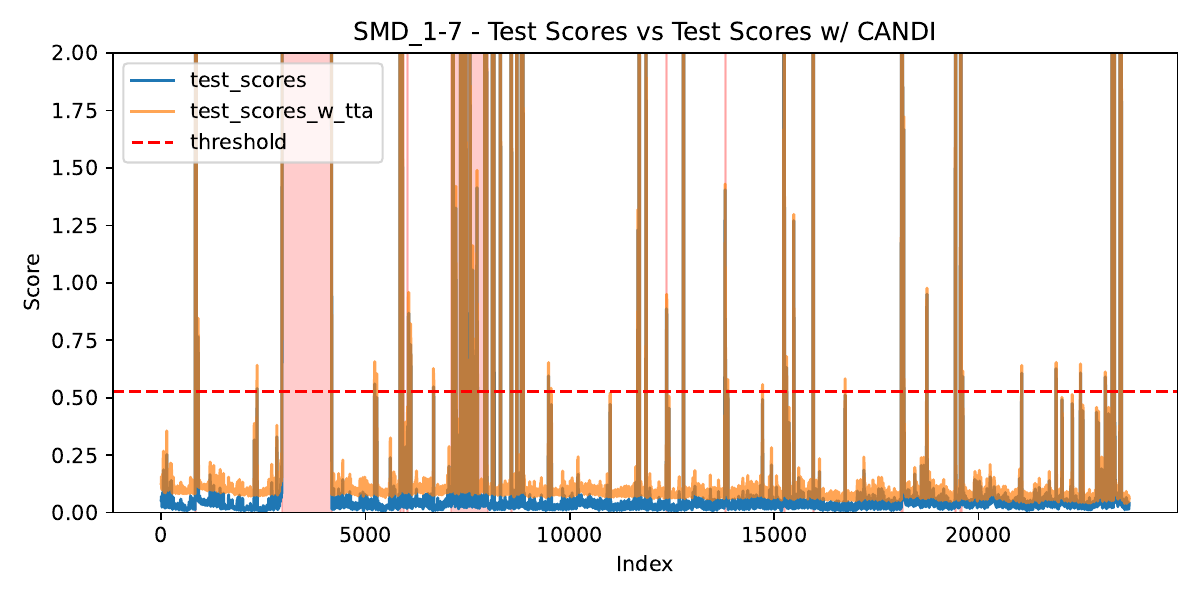}
        \label{fig:scores_smd_1-7}
    \end{minipage}
    
    \vspace{-0.75em}

    \begin{minipage}[b]{0.49\textwidth}
        \centering
        \includegraphics[width=\linewidth]{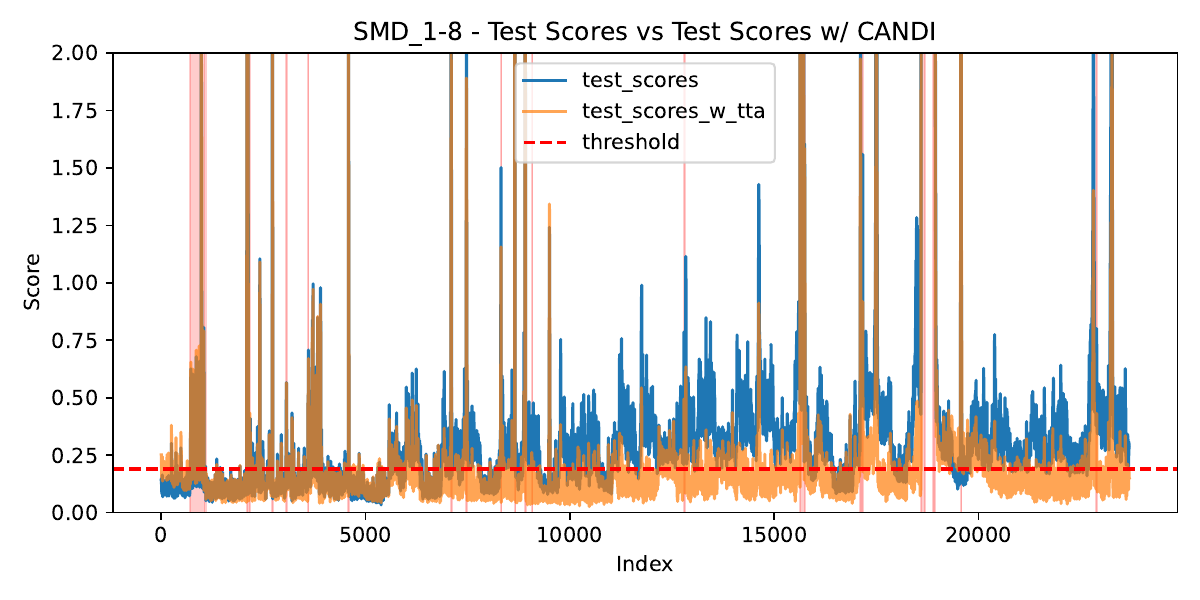}
        \label{fig:scores_smd_1_8}
    \end{minipage}
    \begin{minipage}[b]{0.49\textwidth}
        \centering
        \includegraphics[width=\linewidth]{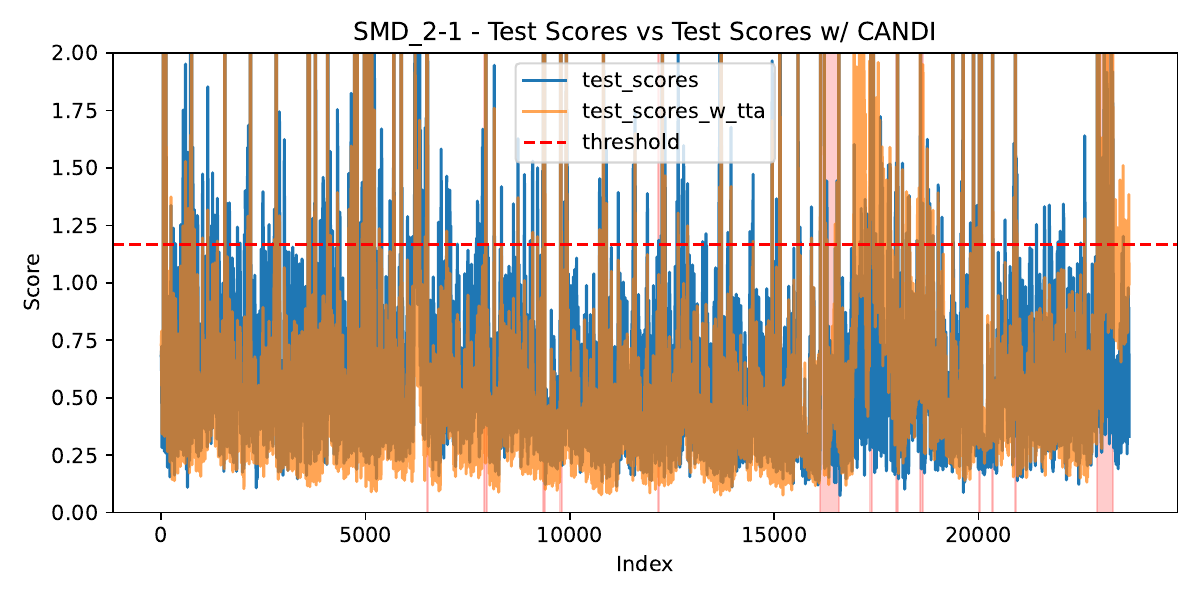}
        \label{fig:scores_smd_2_1}
    \end{minipage}

    \vspace{-0.75em}

    \begin{minipage}[b]{0.49\textwidth}
        \centering
        \includegraphics[width=\linewidth]{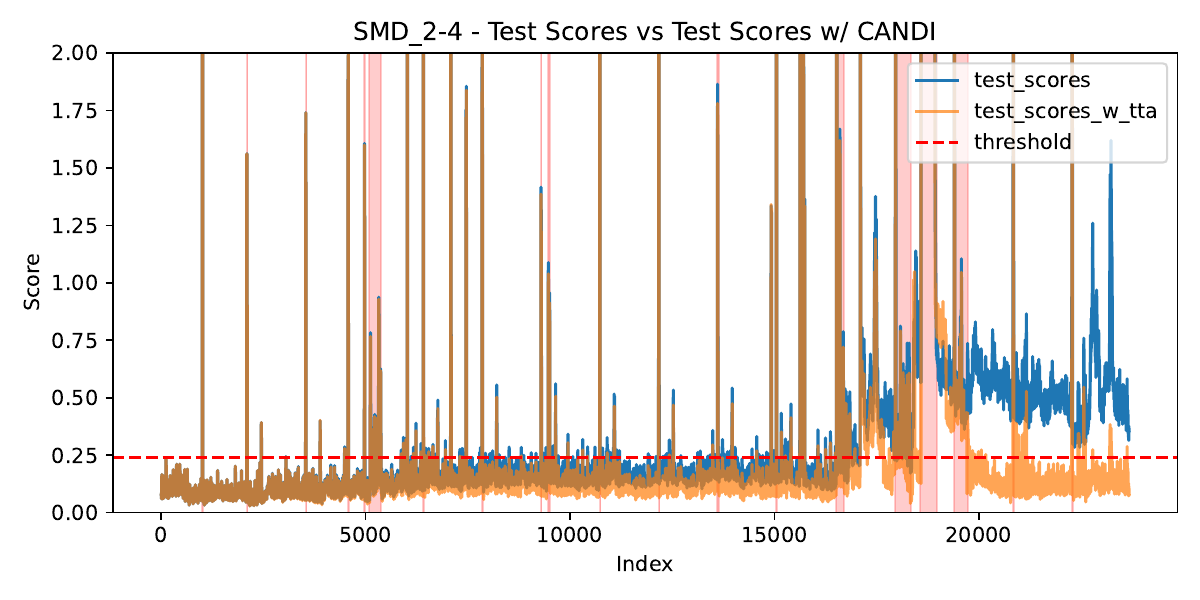}
        \label{fig:scores_smd_2_4}
    \end{minipage}
    \begin{minipage}[b]{0.49\textwidth}
        \centering
        \includegraphics[width=\linewidth]{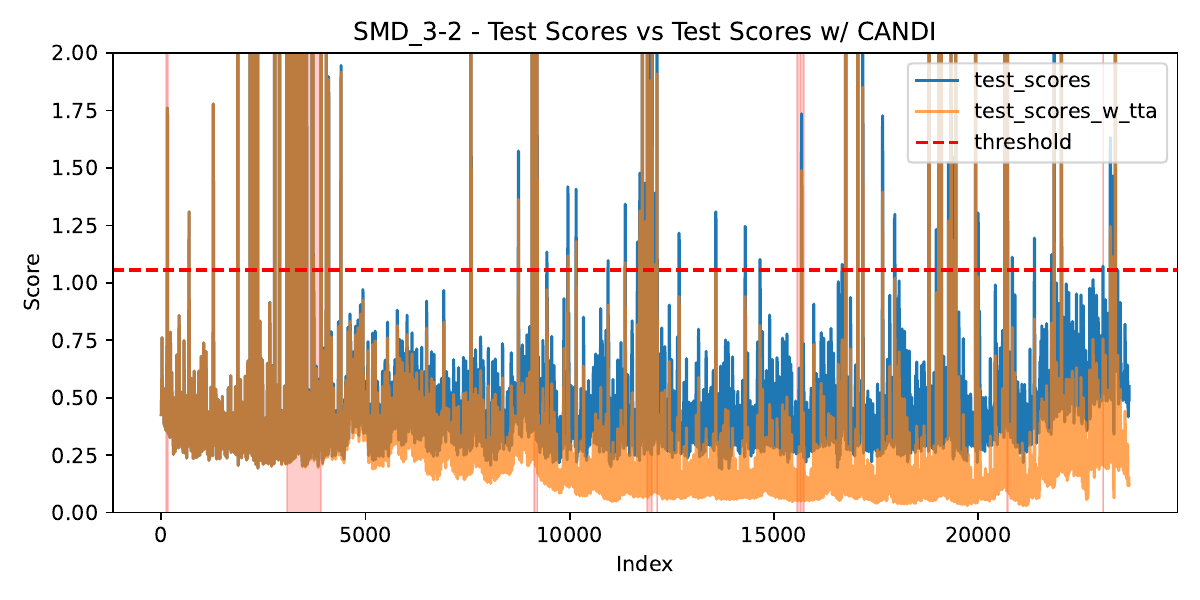}
        \label{fig:scores_smd_3_2}
    \end{minipage}

    \caption{Comparison of test scores with and without CANDI ($\alpha = 5.0$); red shaded regions indicate true anomalies. CANDI lowers scores for normal samples, reducing false positives.
}
    \label{fig:scores_example}
\end{figure*}
Figure~\ref{fig:scores_example} compares the test scores with and without the application of CANDI. The threshold was set with $\alpha = 5.0$. Red shaded regions indicate the true anomaly intervals. When CANDI is applied, the test scores for normal samples tend to be lower compared to the case without CANDI, resulting in fewer false positives.

\section{Limitations and Future Work}
CANDI addresses key challenges in TTA for MTSAD, but several directions remain open for future work.
First, we follow the standard assumption in unsupervised multivariate time-series anomaly detection, where training data is presumed to contain only normal patterns. However, real-world scenarios may involve contaminated or noisy data even in the training data. Developing mechanisms to detect and handle such cases would further enhance robustness. 
We also focus on adapting model parameters, not detection thresholds. Under distribution shift, threshold adjustment may be necessary to control false positives or maintain sensitivity.
CANDI minimizes the risk of adapting to anomalies by curating reliable adaptation samples through false positive mining, rather than indiscriminately updating on all test inputs. This selective strategy helps avoid learning from anomalous patterns that could degrade performance. However, exploring failure recovery mechanisms and strategies to enhance resilience under extreme distribution shifts would be valuable directions for future work. We see these directions as promising steps toward more resilient and versatile TTA systems for MTSAD.

\end{document}